\documentclass[10pt,twocolumn,letterpaper]{article}

\usepackage{cvpr}              
\usepackage{adjustbox,multirow,pifont,float}
\usepackage{amsmath, amssymb, booktabs, amsfonts, nicefrac, microtype, bbding, multirow, tabularx, subcaption, pifont, enumitem, newtxtext, xspace, appendix, array, etoolbox, adjustbox, tikz}
\usepackage{fontawesome}
\usepackage[accsupp]{axessibility}

\definecolor{cvprblue}{rgb}{0.21,0.49,0.74}
\usepackage[pagebackref,breaklinks,colorlinks,allcolors=cvprblue]{hyperref}

\def\paperID{11807} 
\def\confName{CVPR}
\def\confYear{2025}

\title{HiFi-Portrait: Zero-shot Identity-preserved Portrait Generation \\ with High-fidelity Multi-face Fusion}

\author{
Yifang Xu\textsuperscript{1*} \quad 
Benxiang Zhai\textsuperscript{1*} \quad 
Yunzhuo Sun\textsuperscript{2} \quad  
Ming Li\textsuperscript{3} \quad 
Yang Li\textsuperscript{1} \quad
Sidan Du\textsuperscript{1$\dagger$}
\\
\normalsize{\textsuperscript{1}Nanjing University} \quad
\normalsize{\textsuperscript{2}Dalian University of Technology} \quad
\normalsize{\textsuperscript{3}Nanjing University of Information Science and Technology} 
\vspace{-0.1cm}
\\
\footnotesize{
    \{xyf, zbx\}@smail.nju.edu.cn, \quad
    sunyunzhuo\@mail.dlut.edu.cn, \quad
    mingli@nuist.edu.cn, \quad
    \{yogo, coff128\}@nju.edu.cn
}
}

\newcommand{\method}{HiFi-Portrait}

\newcommand\blfootnote[1]{%
  \begingroup
  \renewcommand\thefootnote{}\footnote{#1}%
  \addtocounter{footnote}{-1}%
  \endgroup
}

\begin{document}
\maketitle
\begin{abstract}
Recent advancements in diffusion-based technologies have made significant strides, particularly in identity-preserved portrait generation (IPG). However, when using multiple reference images from the same ID, existing methods typically produce lower-fidelity portraits and struggle to customize face attributes precisely. To address these issues, this paper presents \textbf{\method}, a \textbf{hi}gh-\textbf{fi}delity method for zero-shot \textbf{portrait} generation. Specifically, we first introduce the face refiner and landmark generator to obtain fine-grained multi-face features and 3D-aware face landmarks. The landmarks include the reference ID and the target attributes. Then, we design HiFi-Net to fuse multi-face features and align them with landmarks, which improves ID fidelity and face control. In addition, we devise an automated pipeline to construct an ID-based dataset for training \method. Extensive experimental results demonstrate that our method surpasses the SOTA approaches in face similarity and controllability. Furthermore, our method is also compatible with previous SDXL-based works. 
\end{abstract}

\blfootnote{\text{*} Equal contributions. $\qquad$ $\dagger$ Corresponding author.}

\vspace{-0.6cm}
\section{Introduction}
\label{sec:intro}

Identity-preserved portrait generation (IPG) has garnered significant attention in the past few years because of its potential applications in image animation, virtual try-ons, and e-commerce advertising. It aims to customize portraits consistent with the reference ID. Early approaches \cite{Drawinginstyles-2022, SofGAN-2022}, constrained by the capabilities of generative model \cite{GAN-2014}, suffered from poor face similarity and image diversity. Thanks to the rapid advancement of large-scale text-image datasets \cite{LAION-5B-2022, CLIP-2021} and text-to-image diffusion models \cite{LDM-2021-SD, Imagen-2022}, some methods fine-tune UNet \cite{Textual-Inversion-2022, DreamBooth-2023} or LoRA \cite{LoRA-2021, FaceChain-2023, EasyPhoto-2023, HyperDreamBooth-2023} to create true-to-reality portraits. However, this fine-tuning typically requires at least 30 minutes per ID, which limits their practicality.

\begin{figure}[t!]
  \centering
  \includegraphics[width=\linewidth]{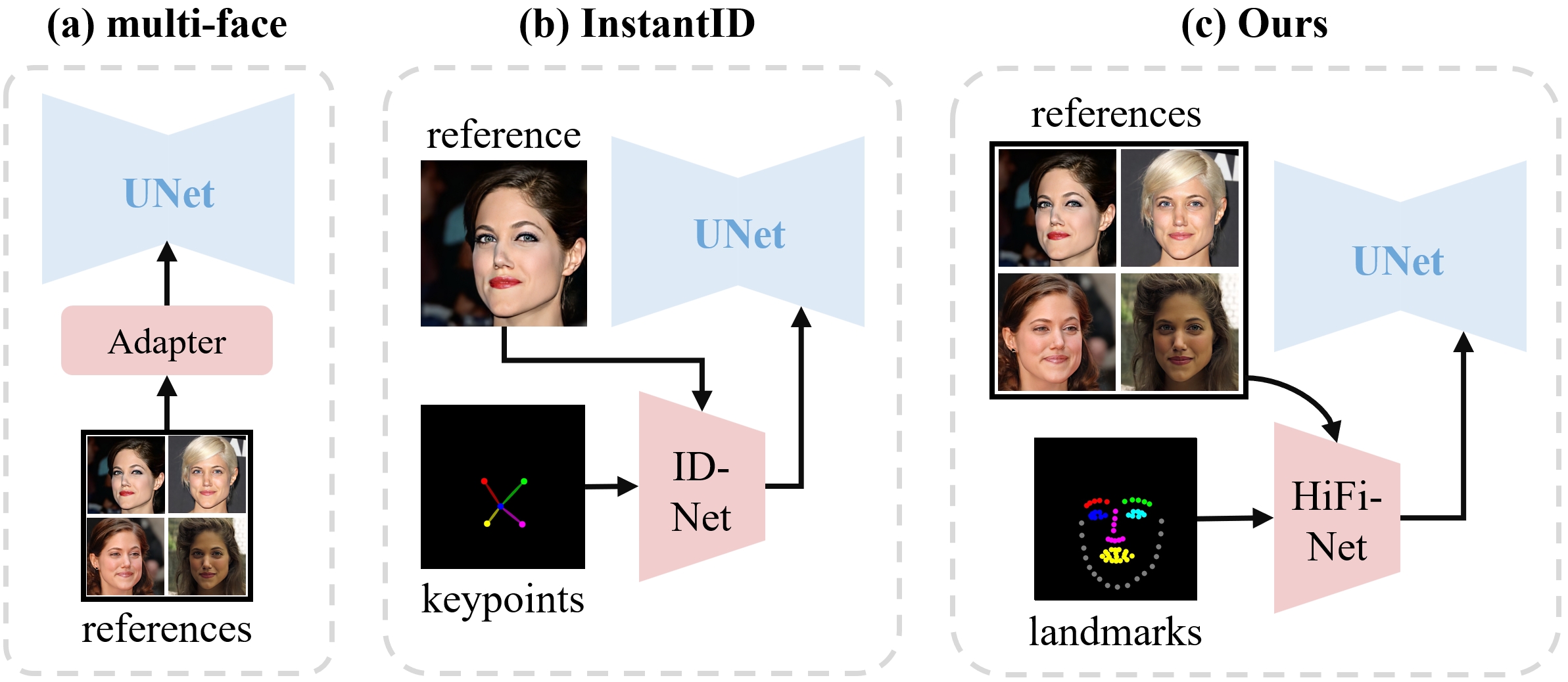}
  \caption{
    (a) Previous multi-face methods lack face location information, obtaining poor results. (b) InstantID only utilizes a single reference image, which limits its performance. (c) Our method fuses multi-faces and aligns them with landmarks via HiFi-Net, achieving high-fidelity ID-preservation. For a more concise presentation, we omit the face encoder.
  }
  \label{fig:high-level}
\vspace{-0.4cm}
\end{figure}

Recently, some studies \cite{IP-Adapter-2023, InstantID-2024, PuLID-2024, FlashFace-2024, IDAdapter-2024, MRNet-2024, MM-Diff-2024, Inv-Adapter, Easyref-2024, VTG-GPT-2023, ConsistentID-2024, Diff-PC-2025, HP3-2025} introduce a zero-shot setup, enabling portrait generation within seconds without fine-tuning during inference. Among them, InstantID \cite{InstantID-2024} is a representative work, as depicted in Fig.\ref{fig:high-level}~(b), utilizing five keypoints (from target face) to guide the reference face in IdentityNet to achieve realistic IPG. Yet, it only considers one reference image during training, limiting its performance. Inspired by face feature fusion \cite{CONAN-2023}, some follow-up works \cite{PhotoMaker-2023, IDAdapter-2024, FlashFace-2024} attempt using Adapter \cite{IP-Adapter-2023} to inject multi-face embeddings from the same ID, as illustrated in Fig.\ref{fig:high-level}~(a). However, these methods tend to generate lower-fidelity portraits due to two factors: (1) Adapter provides weaker control compared to IdentityNet \cite{InstantID-2024}, and (2) the absence of facial positional guidance during multi-face fusion, resulting in inadequate multi-face alignment. In addition, all the above approaches cannot precisely control facial expressions and postures by relying solely on text prompts.

\begin{figure}[t!]
  \centering
  \includegraphics[width=\linewidth]{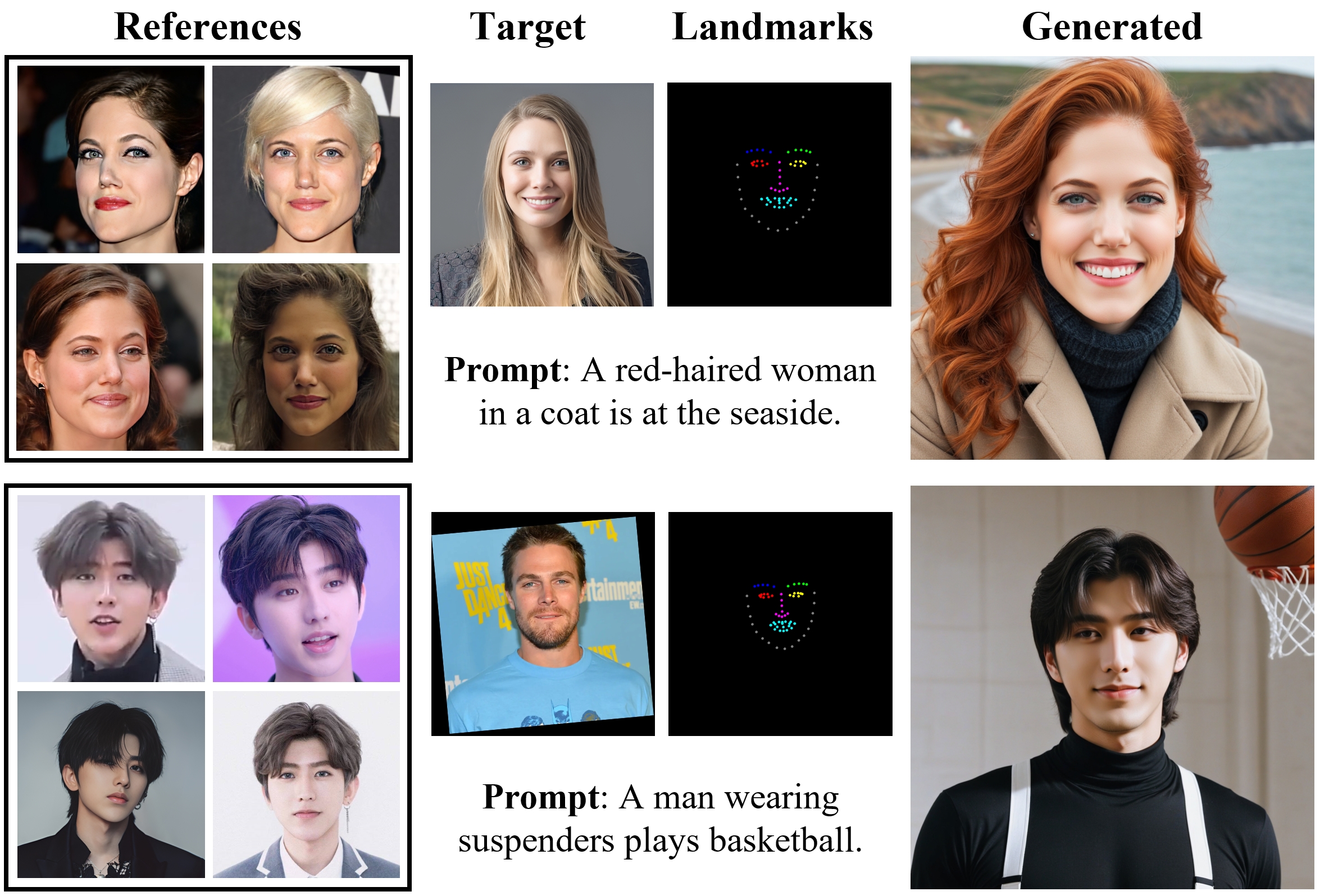}
  \caption{
    Given one or multiple reference images from the same person, \textbf{\method}\space aims to generate high-fidelity portraits preserving the reference ID, where facial expressions and poses follow the target image, while other attributes (such as image background) are personalized via text prompt. 
  }
  \label{fig:2}
\end{figure}

To address the aforementioned challenges, this paper proposes \method\space for zero-shot IPG, a high-fidelity framework incorporating multi-face fusion, as illustrated in Fig.~\ref{fig:2}~and~\ref{fig:method}. To preserve more fine-grained ID information, we first design the face refiner to encode local and global facial features, generating multi-face features. Next, we introduce HiFi-Net to integrate these multi-face features and align them with 3D-aware face landmarks, thus enhancing ID similarity. The landmarks consist of the reference face ID and the expression and pose customized by the target face. Besides, we devise a data collection pipeline to build a diverse ID-oriented dataset, where each ID contains multiple portraits with different attributes and captions. Using this type of data ensures that the model adheres to the guidance of landmarks and text prompts, avoiding direct replication of the reference face. In summary, our primary contributions include:

\begin{itemize}
\item 
We propose \textbf{\method}, a tuning-free method for IPG with high-fidelity, which utilizes HiFi-Net to fuse multi-face features and align them with the landmarks.

\item 
We collect a high-quality \textbf{ID-based dataset} consisting of 34k IDs and 960k text-portrait pairs, designed explicitly for training \method.

\item 
\textbf{High-fidelity.} 
Qualitative and quantitative experiments demonstrate that our method outperforms the SOTA approaches in face similarity and controllability.

\item 
\textbf{Compatibility.} 
Our method is compatible with existing SDXL-based works, achieving high-quality multi-style generation.

\end{itemize}

\section{Related Work}
\label{sec:related_work}

\subsection{Text-to-Image Diffusion Models.} 
Diffusion models (DMs) \cite{DDPM-2020, DDIM-2020, Improved-Diffusion-2021, WAM-Flow} recently made significant advancements in image synthesis, which has attracted considerable attention. Compared to autoregressive models \cite{Autoregressive-2011} and generative adversarial networks \cite{DCGAN-2015, GAN-2014}, text-to-image DMs \cite{VQ-Diffusion-2021, Imagen-2022, unCLIP-2022-DALL-E-2, Parti-2022} excel in generating diverse and high-quality images based on user-provided text prompts. A representative example is Stable Diffusion (SD) \cite{LDM-2021-SD}, which utilizes the CLIP \cite{CLIP-2021} text encoder to obtain text embeddings and then guide the denoising process through cross-attention mechanisms. Moreover, it performs denoising in latent space, thus significantly saving computational resources. Building on this, SDXL \cite{SDXL-2023} enhances textual control by expanding the U-Net architecture and incorporating an additional text encoder. 

\subsection{ID-Preserved Portrait Generation.} 
The goal of IPG is to generate portraits that are consistent with the reference ID, while being personalized based on given conditions (e.g., textual prompt, depth map) \cite{Textual-Inversion-2022, DreamBooth-2023, EasyPhoto-2023, PFANet-2021}. Initially, Textual Inversion \cite{Textual-Inversion-2022} achieves this by training text embeddings with multiple images of the same ID. DreamBooth \cite{DreamBooth-2023} accomplishes this through fine-tuning UNet. To reduce trainable parameters, some works train low-rank adaptation \cite{LoRA-2021, HyperDreamBooth-2023, EasyPhoto-2023} to preserve facial ID. However, all the above methods require at least 30 minutes of individual training for each ID during inference, which is highly time-consuming and the applicability of IPG.

\subsection{Zero-shot ID-Preserved Portrait Generation.}
Some recent studies \cite{IP-Adapter-2023, InstantID-2024, PuLID-2024, ConsistentID-2024, IDAdapter-2024, GPTSee-2023, Moment-GPT-2025, MM-Diff-2024, Inv-Adapter, MH-DETR-2024} introduce zero-shot IPG, which leverages large-scale text-image datasets to train models. During inference, these models customize portraits within seconds using reference images without additional fine-tuning. For instance, IP-Adapter-FaceID \cite{IDAdapter-2024} trains Adapter (consisting of projection layer and cross-attention) to inject face embeddings and maintain ID consistency. However, the limited parameters in Adapter restrict the ID fidelity of results. To address this, the existing SOTA method, InstantID \cite{InstantID-2024}, utilizes five face keypoints to guide the face embeddings in IdentityNet. IdentityNet's architecture is similar to that of ControlNet \cite{ControlNet-2023}, but its inputs and conditions are the keypoints and face embeddings, respectively. However, InstantID only utilizes one reference face, which limits its performance.

To enhance facial similarity, subsequent zero-shot approaches \cite{PhotoMaker-2023, IDAdapter-2024, FlashFace-2024} employ the "multi-face" strategy, which uses multiple reference images of the same individual for training. PhotoMaker \cite{PhotoMaker-2023} and IDAdapter \cite{IDAdapter-2024} incorporate multi-face embeddings through cross-attention. FlashFace extends this approach by additionally training a face encoder to strengthen ID control. Nevertheless, these multi-face methods ignore the combination of local and global facial features. Moreover, they lack guidance regarding facial position information, leading to low fidelity of generated results. To tackle these challenges, this paper proposes \method, which fuses multi-face features guided by the landmarks, thus enhancing ID fidelity. In addition, we collect an ID-driven dataset to train our model, thereby further improving ID similarity.


\begin{figure*}[t!]
  \centering
  \includegraphics[width=\linewidth]{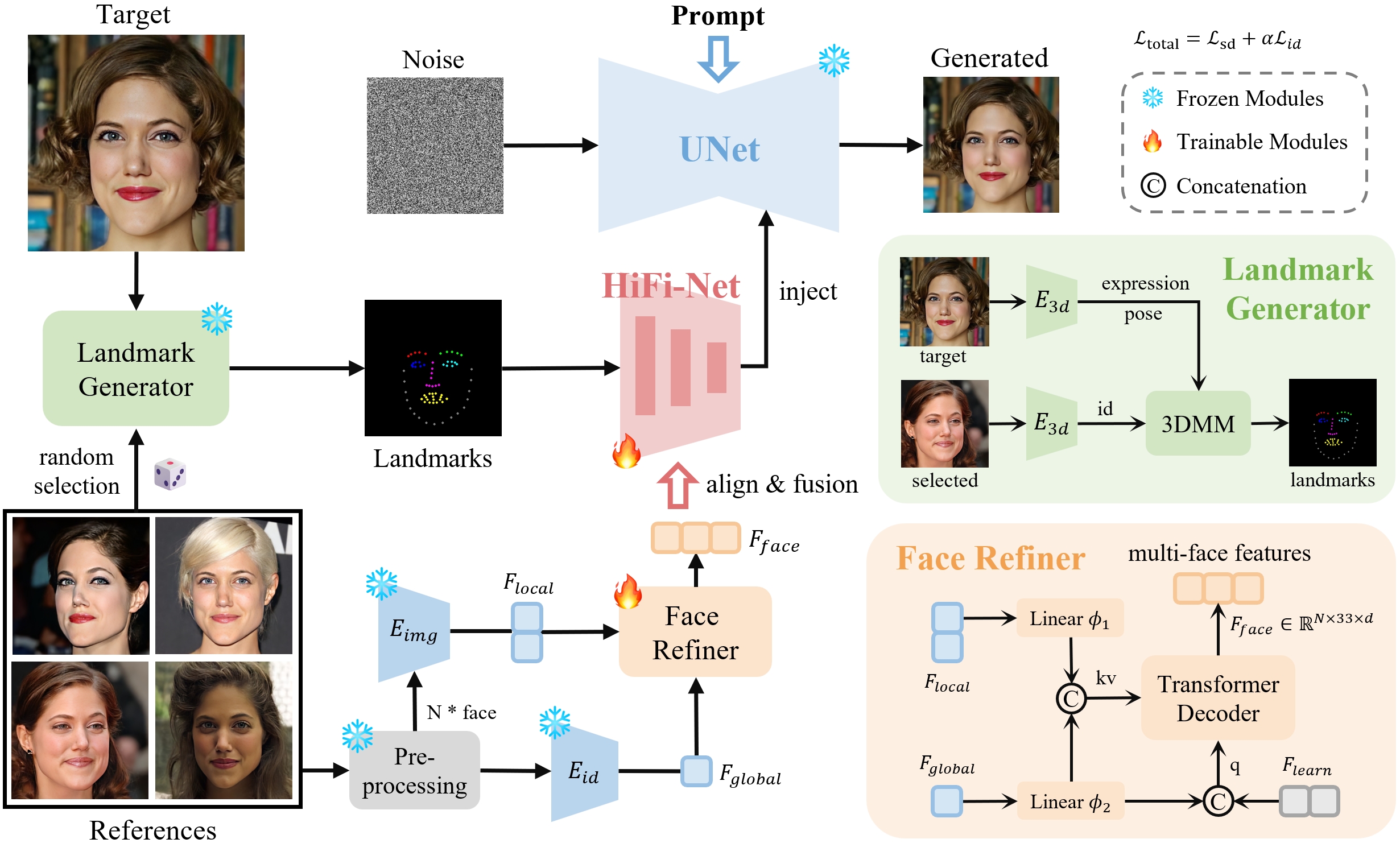}
  \caption{
    \textbf{The overall framework of \method}. It consists of three key modules: (1) the face refiner merges local and global facial features to retain ID details. (2) the landmark generator produces 3D-aware landmarks, providing accurate guidance for facial control. (3) HiFi-Net fuses multi-face features and aligns them with landmarks, improving ID fidelity and controllability.
  }
  \label{fig:method}
\end{figure*}

\section{Method}
\label{sec:method}
In this section, we first review the principles of text-to-image diffusion models (Sec.~\ref{subsec:Preliminary}). Next, Sec.~\ref{subsec:method} elaborates on the module and design motivation of \method. Sec.~\ref{subsec:train_infer} introduces the training and inference processes. Finally, Sec.~\ref{subsec:pipeline} details the pipeline for constructing an ID-oriented dataset.

\subsection{Preliminary}
\label{subsec:Preliminary}
We build our method on SDXL \cite{SDXL-2023}, which enhances efficiency by projecting data into low-dimensional latent spaces. Specifically, SDXL first employs VAE \cite{VAE-2014} encoder $\mathcal{E}$ to compress the given image $x_{0} \in \mathbb{R}^{H \times W \times 3}$ into a latent code $z = \mathcal{E} (x_{0}) \in \mathbb{R}^{\frac{H}{8} \times \frac{W}{8} \times 4}$. Subsequently, the temporal-conditional UNet denoiser $\epsilon_{\theta}$ estimates the noise added at the different timestep. The training objective can be defined as follows:
\begin{align}
    \mathcal{L} = \mathbb{E}_{z_t, t, c, \epsilon \sim \mathcal{N}(0,1)} [|| \epsilon - \epsilon_{\theta}(z_t, t, c)||_{2}]
\end{align}
where $z_{t}$ represents noisy latent at the $t$-th timestep. $c$ denotes the text embeddings encoded by the CLIP text encoder. $\epsilon$ is ground-truth noise sampled from a standard Gaussian distribution.

\subsection{\method}
\label{subsec:method}

\paragraph{Overview.} 
Given $N$ reference images $I_{ref} \in \mathbb{R}^{N \times H \times W \times 3}$ of the same individual, \method\space aims to generate high-fidelity portrait $\hat{x}_{0} \in \mathbb{R}^{H \times W \times 3}$ preserving the reference ID, where facial expressions and poses follow the target image $I_{tgt} \in \mathbb{R}^{H \times W \times 3}$, while other attributes (such as image style and background) are customized via text prompt. 

Fig.~\ref{fig:method} depicts the overall framework of \method. It comprises three key components: (1) face refiner, which integrates local and global facial features to preserve detailed ID information. (2) landmark generator, creating 3D prior landmarks for accurate facial control. (3) HiFi-Net, which aligns and fuses multi-face features to enhance ID-fidelity.







\vspace{-0.45cm}
\paragraph{Pre-processing.}
For the given reference images $I_{ref}$, we first apply PP-Matting \cite{PP-matting} for human matting to remove the cluttered background in $I_{ref}$. Next, we employ the face detection model $E_{det}$ \cite{RetinaFace-2020} to obtain the highest-scoring faces $I_{face} \in \mathbb{R}^{N \times h \times w \times 3}$.

\vspace{-0.35cm}
\paragraph{Face encoder.}
We then encode the reference faces $I_{face}$ into facial features at different granularities. Specifically, we utilize the CLIP image encoder $E_{img}$ to derive local facial features $F_{local} = E_{img}(I_{face}) \in \mathbb{R}^{N \times 257 \times d_1}$. Subsequently, we use the identity extractor $E_{id}$ \cite{ArcFace-2019}, pre-trained on face recognition and person re-identification, to obtain global facial features $F_{global} = E_{id}(I_{face}) \in \mathbb{R}^{N \times 1 \times d_2}$.

\vspace{-0.3cm}
\paragraph{Face refiner.}
Since $F_{local}$ contains fine-grained facial appearance, $F_{global}$ contains key facial ID, they are complementary. A simple approach is concatenating and inputting them into HiFi-Net. However, prior research \cite{InstantID-2024} indicates that using $F_{local}$ can result in the direct pasting of blurred reference faces onto the generated images, thereby reducing face similarity. Thus, we design the face refiner to address this issue. 

Specially, we first employ linear layers $\phi_{1}$ and $\phi_{2}$ for mapping $F_{local}$ and $F_{global}$ to the same dimension $d$, where $F_{local}^{'} = \phi_{2}(F_{local}) \in \mathbb{R}^{N \times 257 \times d}$ and $F_{global}^{'} = \phi_{1}(F_{global}) \in \mathbb{R}^{N \times 1 \times d}$. Then, we leverage the transformer decoder \cite{Transformer} with learnable face tokens $F_{learn} \in \mathbb{R}^{N \times 32 \times d}$ to extract ID details from $F_{local}^{'}$ and $F_{global}^{'}$, thereby generating multi-face features $F_{face} \in \mathbb{R}^{N \times 33 \times d}$. Here, we concatenate $F_{global}$ with $F_{learn}$ as \textit{query}, and concatenate $F_{global}$ with $F_{local}$ as \textit{key} (\textit{value}). The above process can be defined as follows:
\begin{gather}
    q = \text{concat}(F_{global}^{'}, F_{learn}) \\
    kv = \text{concat}(F_{global}^{'}, F_{local}^{'}) \\
    F_{face} = \text{Decoder}(q, kv)
\end{gather}

\vspace{-0.55cm}
\paragraph{Landmark generator.}
As shown in Fig.~\ref{fig:method}, we first utilize the 3D reconstruction network, DECA \cite{DECA-2021}, to extract facial ID, expression, and pose vectors from reference and target images. Subsequently, we combine the reference ID with target expression and pose vectors, then render a new 3D face using the 3D Morphable Model (3DMM) fitting algorithm \cite{3DMM-1999}. Finally, we project the 3D face onto the 2D plane, resulting in 3D-aware face landmarks $I_{lmk} \in \mathbb{R}^{H \times W \times 3}$.

\vspace{-0.3cm}
\paragraph{HiFi-Net.}
To enhance ID fidelity, we input conditions $I_{lmk}$ and $F_{face}$ into HiFi-Net, a network architecture derived from ControlNet \cite{ControlNet-2023}. Specifically, the VAE first embeds $I_{lmk}$ into latent space, yielding latent code $z_{lmk} \in \mathbb{R}^{\frac{H}{8} \times \frac{W}{8} \times 4}$. Subsequently, $z_{lmk}$ guides the fusion and alignment of $F_{face}$ into the latent spaces, facilitated by the cross-attention mechanism within HiFi-Net. This process is mathematically represented as follows:
\begin{gather}
    \text{Attention}(Q, K, V) = \text{softmax} \left(\frac{QK^{T}}{\sqrt{d_k}}\right) V
\end{gather}
where $Q=\phi_{q}(z_{lmk})$, $K=\phi_{k}(F_{face}^{'})$, and $V=\phi_{v}(F_{face}^{'})$ represent the \textit{query}, \textit{key}, and \textit{value} respectively. $F_{face}^{'} \in \mathbb{R}^{(N \times 33) \times d}$ denotes the flattened $F_{face}$. The functions $\phi_{q}$, $\phi_{k}$, and $\phi_{v}$ are linear layers. Finally, HiFi-Net yields high-fidelity identity features $c_{id}$, which are injected into the UNet denoiser to preserve facial identity effectively. The architecture of HiFi-Net can be found in the \textbf{\textit{Appendix}}.

\subsection{Model Training and Inference}
\label{subsec:train_infer}

\paragraph{Training.}
During training, we randomly select one image from $I_{ref}$, and feed it into the landmark generator to obtain the reference ID. We only train the parameters of the face refiner and HiFi-Net, while the other modules remain frozen. The optimization objective of SD is as follows: 
\begin{equation}
    \mathcal{L}_{sd} = || \epsilon - \epsilon_{\theta}(z_t, t, c, c_{id})||_{2}
\end{equation}
In addtion, we follow previous works \cite{ArcFace-2019, IDAdapter-2024}, employing the face ID loss $\mathcal{L}_{id}$ to improving ID fidelity. 
%
\begin{equation}
\label{eq:id_loss}
    \mathcal{L}_{id} = E_{det}(\hat{x}_{0}) \cdot [1 - \text{cos}(F_{id}, E_{id}(\hat{x}_{0}))]
\end{equation}
Here, $\text{cos}$ is the cosine similarity. $F_{id} = \text{avg}[E_{id}(I_{face})] \in \mathbb{R}^{1 \times d_{2}}$ represents the average reference ID features. $\hat{x}_{0}$ denotes the generated denoised image. We use $E_{det}$ to prevent the unclear face in $\hat{x}_{0}$ from misleading the model. When a face is detected in $\hat{x}_{0}$, $E_{det}(\hat{x}_{0}) = 1$; otherwise $E_{det}(\hat{x}_{0}) = 0$. Finally, the total loss with $\alpha=0.1$ is as follows:
\begin{equation}
\label{eq:total_loss}
    \mathcal{L}_{total} = \mathcal{L}_{sd} + \alpha \mathcal{L}_{id}
\end{equation}

\vspace{-0.55cm}
\paragraph{Inference.}
During inference, we compute the face score $S_{ref} \in \mathbb{R}^{N}$ for the given $N$ reference images $I_{ref}$ (see Eq.~\ref{eq:infer}). Subsequently, we select the highest-scoring reference image into the landmark generator to get reference
\begin{equation}
\label{eq:infer}
    S_{ref} = S_{q}(I_{ref}) + \lambda S_{a}(I_{ref}, I_{tgt})
\end{equation}
where $S_{q}$ represents the face quality score estimated by ModelScopeFQA \cite{ModelScopeFQA}. $S_{q}$ is higher for images with greater facial clarity and minimal obstructions. $S_{a}$ is the face angle score calculated by RetinaFace \cite{RetinaFace-2020}. Using $S_{a}$ aims to select the images that are more consistent with the angle of target image $I_{tgt}$, which reduces the loss of ID information in 3DMM. $\lambda$ is set to 0.5, balancing the influence of quality and angle in the face score.



\begin{figure*}[t!]
  \centering
  \includegraphics[width=\linewidth]{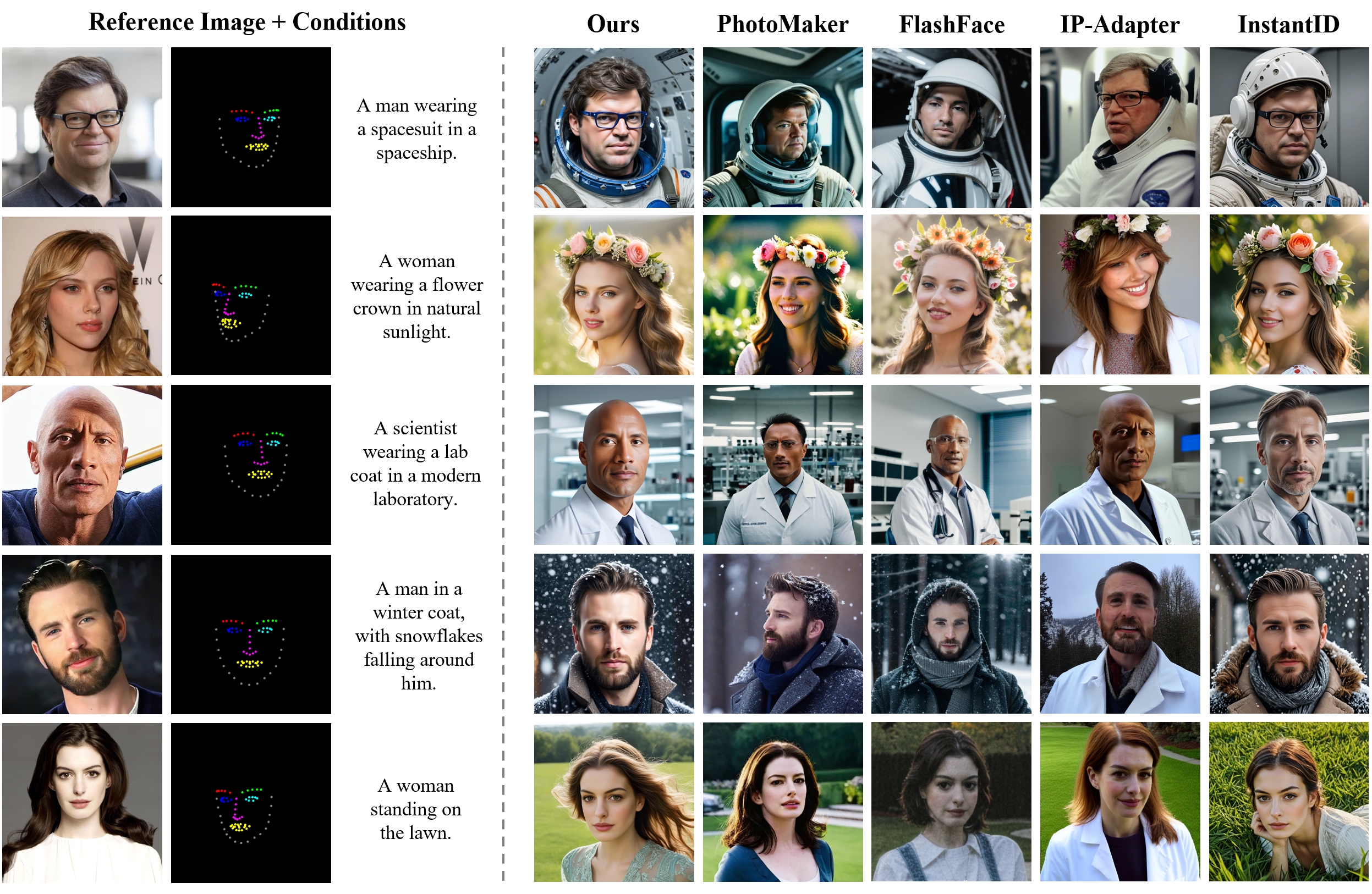}
  \caption{
    \textbf{Qualitative results}. We compare \method\space with four of the SOTA open-source techniques: PhotoMaker \cite{PhotoMaker-2023}, FlashFace \cite{FlashFace-2024}, IP-Adapter \cite{IP-Adapter-2023}, and InstantID \cite{InstantID-2024}. All methods use 4 reference portraits from the same ID. We observe that our framework achieves high-fidelity portrait generation with ID preservation, with precise control over image attributes facilitated by conditions (landmarks and text prompts).
  }
  \label{fig:4}
\vspace{-0.3cm}
\end{figure*}

\subsection{ID-based Dataset Construction Pipeline}
\label{subsec:pipeline}
To enhance ID fidelity, \method\space requires multiple images from the same ID for training. However, existing datasets in multi-face IPG methods \cite{PhotoMaker-2023, FlashFace-2024} are not yet open-source. To address this, we devise a high-quality pipeline for creating an ID-based dataset. This dataset includes an average of 28 text-image pairs per ID, with images featuring diverse expressions, poses, hairstyles, and other attributes. We believe that this dataset will advance future ID-based (multi-face) research.

\vspace{-0.35cm}
\paragraph{Data collection.}
We utilize the open-source datasets VGGFace2-HD \cite{dataset-VGGFace2-2018, SimSwapPP-VGGFaceHD-2023} and IMDb-Face \cite{dataset-IMDb-Face-2018} for our research. For IMDb-Face, we download raw images from all valid source urls. To enhance ID diversity, we additionally gather a list of celebrities from not included in IMDb-Face. Based on this list, we collect approximately 100 images per name via search engines.

\vspace{-0.5cm}
\paragraph{Face detection.}
We initially apply RetinaFace to detect face and obtain bounding boxes, deleting images where the number of faces is not equal to 1. Subsequently, we crop images to larger squares according to the bounding boxes, ensuring the face occupies more than 10\% of the cropped area. In addition, images smaller than 256$\times$256 are discarded, and all remaining images are resized to 1024$\times$1024.

\vspace{-0.4cm}
\paragraph{Data cleansing.}
To improve the dataset quality, we perform the following steps to eliminate noisy images. First, we employ ModelScopeFQA to assess face quality scores, removing images with scores below 0.45. Next, we use PaddleOCR \cite{PaddleOCR} to recognize and delete images with watermarks on faces. To minimize images with occluded faces, we leverage BLIP-2 \cite{BLIP2-2023} to generate facial descriptions, filtering out images that contain foreign objects (e.g., hand, microphone) in the descriptions. Finally, GFPGAN \cite{GFP-GAN-2021} is used for face restoration.

\vspace{-0.4cm}
\paragraph{Face verification.}
To identify images belonging to the current ID group, we extract face embeddings via InsightFace \cite{ArcFace-2019}. Then, we adopt k-means clustering algorithm to choose the central embedding of the largest cluster as the target embedding. Finally, we only save images that similarity with target embedding higher than 0.7, effectively removing images from other IDs.

\begin{table*}[t!]
\caption{
    \textbf{Quantitative comparison} of \method\space and state-of-the-art methods. All metrics are in percentages (\%). The best results are highlighted in \textbf{bold}.
}
\label{tab:qualitative}

\centering
\begin{adjustbox}{max width=0.95\linewidth}
\begin{tabular}{@{}ccccccccccc@{}}
\toprule
\multirow{2}{*}{\textbf{Method}} & \multicolumn{5}{c}{\textbf{WebFace-100}} & \multicolumn{5}{c}{\textbf{InternetFace-300}} \\ 
\cmidrule(l){2-6} \cmidrule(l){7-11} 
& $Sim_{f}\uparrow$ & $CLIP_{f}\uparrow$ & $Exp\downarrow$ & $Pose\downarrow$ & $FID\downarrow$ & $Sim_{f}\uparrow$ & $CLIP_{f}\uparrow$ & $Exp\downarrow$ & $Pose\downarrow$ & $FID\downarrow$ \\ 
\midrule
PhotoMaker \cite{PhotoMaker-2023} & 46.6 & 48.0 & 23.1 & 8.3 & 121.8 & 37.1 & 44.3 & 24.7 & 9.5 & 189.8 \\
ConsistentID \cite{ConsistentID-2024} & 61.4 & 58.5 & 22.7 & 8.0 & 118.3 & 54.2 & 56.1 & 24.3 & 8.7 & 182.4 \\
FlashFace \cite{FlashFace-2024} & 64.3 & 70.2 & 20.9 & 7.9 & 123.4 & 60.4 & \textbf{71.4} & 22.5 & 8.3 & 173.5 \\
IP-Adapter \cite{IP-Adapter-2023} & 64.1 & 65.6 & 20.3 & 8.5 & 120.7 & 62.3 & 63.2 & 23.1 & 8.2 & 176.2 \\
InstantID \cite{InstantID-2024} & 72.9 & 68.2 & 18.8 & 6.1 & 114.4 & 69.7 & 67.8 & 20.4 & 6.7 & \textbf{158.1} \\
\textbf{HiFi-Portrait} (Ours) & \textbf{77.7} & \textbf{72.4} & \textbf{14.7} & \textbf{5.4} & \textbf{108.5} & \textbf{73.1} & 71.1 & \textbf{16.9} & \textbf{6.2} & 163.4 \\
\bottomrule
\end{tabular}
\end{adjustbox}
\vspace{-0.2cm}
\end{table*}

\vspace{-0.4cm}
\paragraph{Image caption.}
Although the landmarks encompass the individual's expressions and poses, to ensure the diversity of the generated images, we prompt Yi-Vision \cite{01ww-2024} for detailed image captions using the following prompt: "\textit{Dear Yi-Vision, please describe this portrait in detail, focusing on the person's appearance, hairstyle, clothing, accessories, and background. Please do not describe facial expressions and poses.}"


\vspace{-0.4cm}
\paragraph{Data statistics.}
Finally, using the proposed pipeline, we create a training dataset comprising 34k IDs and 960k images, with an average of 28 images per ID. For more detailed dataset statistics, please read the \textbf{\textit{appendix}}.



\section{Experiments}
\label{sec:experiments}


\begin{figure}[t!]
  \centering
  \includegraphics[width=\linewidth]{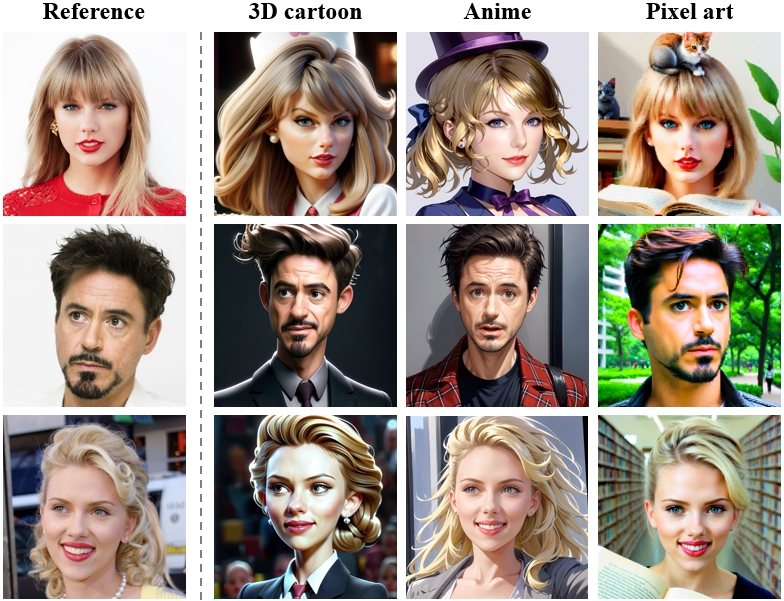}
  \caption{
    \textbf{Compatibility}. Our method is compatible with models of different styles, which can generate diverse and high-quality images with high-fidelity ID preservation.
  }
  \label{fig:application}
  \vspace{-0.3cm}
\end{figure}

\vspace{-0.1cm}
\subsection{Experimental Settings}
\label{subsec:settings}

\vspace{-0.1cm}
\paragraph{Implementation details.}
Our framework is based on SDXL-base-1.0 \cite{SDXL-2023, SDXL-based-1-2024} with the dimension $d = 2048$. $d_{1}=1280$, $d_{2}=512$. Correspondingly, the training images are resized to $1024 \times 1024$. The number of transformer decoder layers in the face refiner is set to 5. For the CLIP encoder, we utilize clip-vit-large-patch14 \cite{CLIP-2021} pretrained by OpenAI. During training, we randomly select 1 image from the same ID group as the target and $N$ images as references, where $N$ is randomly set between 1 and 4. An empty text prompt is also used with 10\% probability to enhance generative performance through classifier-free guidance. We employ the Adam optimizer with a constant learning rate of 2e-5. The model is trained using 4 NVIDIA A800 GPUs over 800k iterations (16 days), with a batch size 24. During inference, we set $N=4$. Use DDIM \cite{DDIM-2020} sampler with 8-steps SDXL-Lightning \cite{SDXL-Lightning-2024}.


\begin{figure}[t!]
  \centering
  \includegraphics[width=\linewidth]{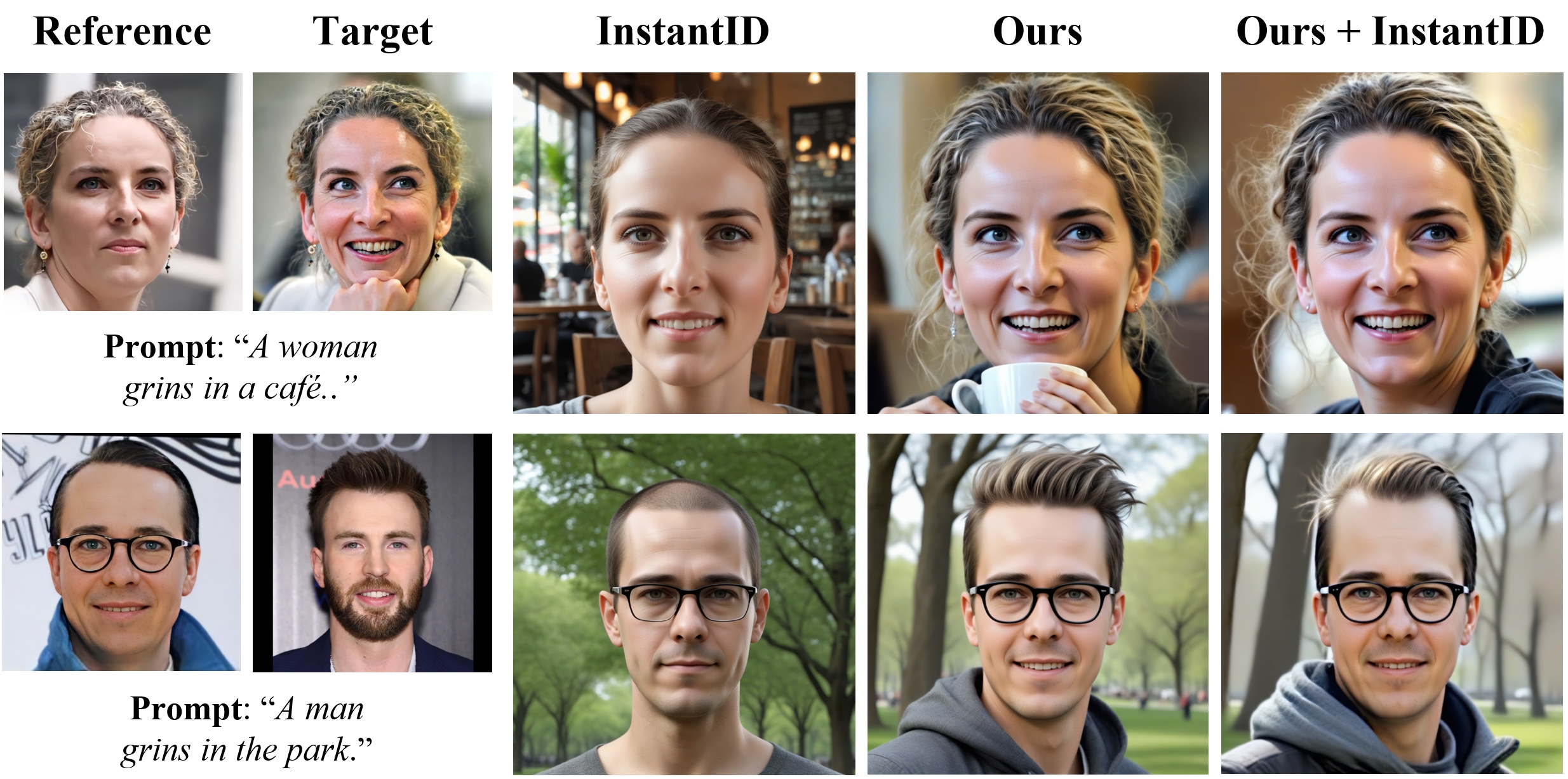}
  \caption{
    \textbf{\method\space + InstantID}. The reference and target image in the top case are from the same ID. The bottom case from the different IDs.
  }
  \label{fig:ours_instantid}
  \vspace{-0.1cm}
\end{figure}

\begin{figure}[t!]
  \centering
  \includegraphics[width=\linewidth]{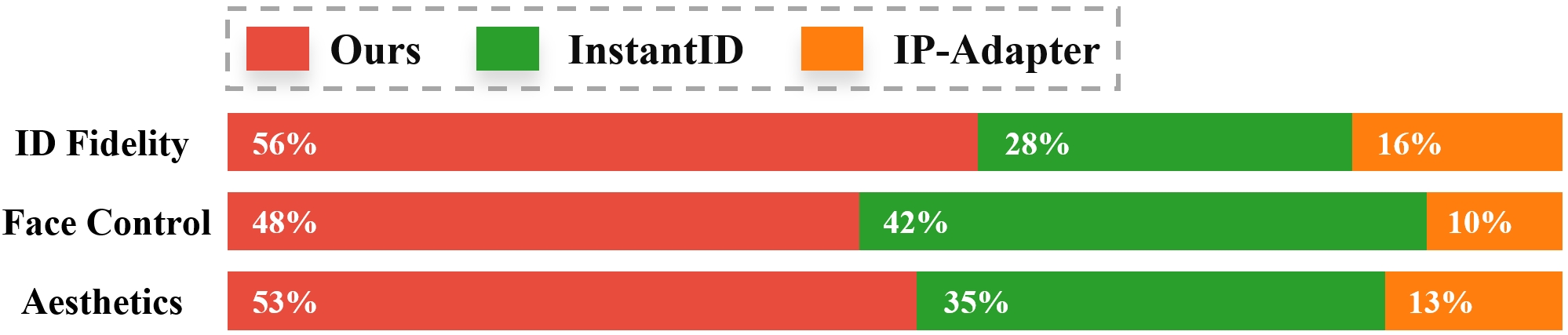}
  \caption{
    \textbf{User study}. Participants consistently prefer \method\space in three aspects: ID fidelity, face controllability, and aesthetics.
  }
  \label{fig:user_study}
  \vspace{-0.3cm}
\end{figure}

\begin{figure*}[t!]
  \centering
  \begin{minipage}{0.75\linewidth}
    \centering
    \includegraphics[width=\linewidth]{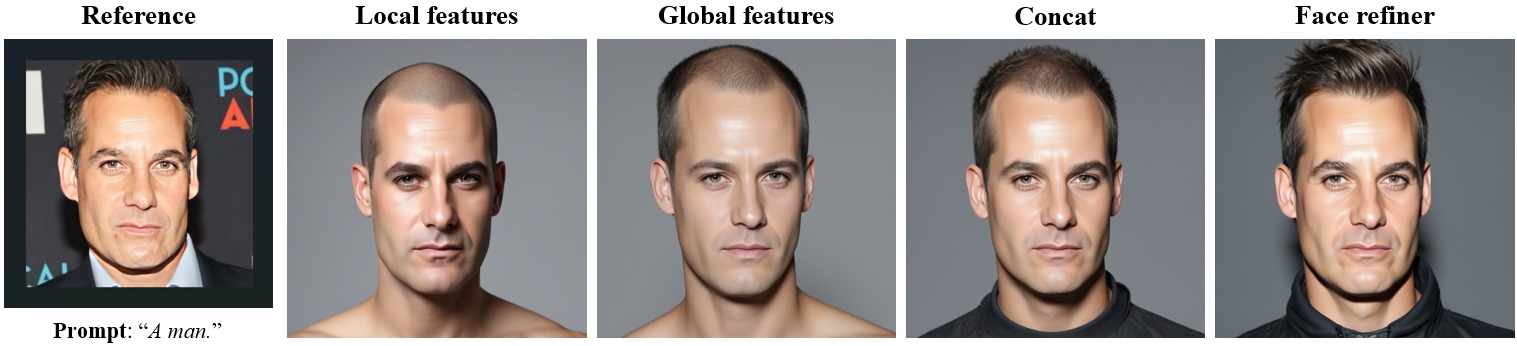}
    \caption{
      \textbf{Impact of face refiner.} Compared with "concat" (directly concatenate local and global features), using the face refiner to obtain multi-face features effectively captures the ID details from local and global features, thereby enhancing ID fidelity.
    }
    \label{fig:face_refiner}
  \end{minipage}
  \hspace{0.005\linewidth}
  \begin{minipage}{0.23\linewidth}
    \captionsetup{type=table} 
    \caption{
        Ablation on face refiner.
    }
    \label{tab:face_refiner}
    \centering
    \begin{adjustbox}{max width=\linewidth}
    \begin{tabular}{@{}ccc@{}}
    \toprule
    \textbf{Method} & $Sim_{f}\uparrow$ & $CLIP_{f}\uparrow$ \\ 
    \midrule
    local features & 62.3 & 55.2 \\
    global features & 63.9 & 56.8 \\
    concat & 64.4 & 57.3 \\
    \textbf{face refiner} & \textbf{65.2} & \textbf{58.9} \\ 
    \bottomrule
    \end{tabular}
    \end{adjustbox}
  \end{minipage}
  \vspace{-0.1cm}
\end{figure*}

\begin{figure*}[t!]
  \centering
  \includegraphics[width=\linewidth]{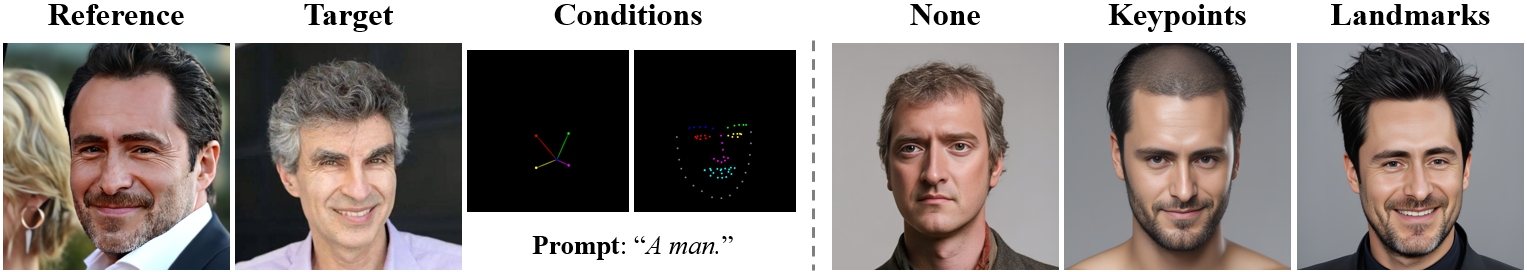}
  \caption{
    \textbf{Effect of HiFi-Net.} "\textit{None}" utilizes Adapter that injects multi-face features through trainable cross-attention layers. "\textit{Keypoints}" maintain HiFi-Net architecture but use face keypoints as input conditions. "\textit{Landmarks}" represent using HiFi-Net. Using landmarks to guide the fusion of multi-face features and aligning them in HiFi-Net achieves high-fidelity results.
  }
  \label{fig:HiFi_Net}
\end{figure*}

\vspace{-0.4cm}
\paragraph{Evaluation datasets.}
To obtain quantitative results, we construct two evaluation datasets. Specifically, we randomly select 100 identities from the WebFace260M \cite{dataset-WebFace260M-2021} dataset, referred to as WebFace-100. We collect 300 ID groups from the Internet that are not used in training, termed InternetFace-300. By leveraging the data processing described in Sec.~\ref{subsec:pipeline}, each ID group comprises 1 target image and 4 reference images. The target image is randomly selected from LAION-Face
\cite{zheng2022general}. We collect 10 diverse text prompts and generate 3 distinct images for each combination (prompt, ID group). In total, we generate 3,000 images for WebFace-100 and 9,000 images for InternetFace-300.

\vspace{-0.4cm}
\paragraph{Evaluation metrics.} 
We employ $Sim_{f}$ and $CLIP_{f}$ to assess ID fidelity, where $Sim_{f}$ ($CLIP_{f}$) represents the average cosine similarity of face (CLIP) embeddings between all reference faces and the generated face. These embeddings are extracted using InsightFace \cite{ArcFace-2019} (CLIP \cite{CLIP-2021} image encoder). To evaluate face controllability, we use $Exp$ and $Pose$, computed as the root mean squared error (RMSE) between the target and generated expression (pose) vectors, estimated by DECA \cite{DECA-2021}. Additionally, we utilize $FID$ \cite{heusel2017gans} to measure the generation quality.


\subsection{Comparisons}
\label{subsec:comparison}

\paragraph{Qualitative results.}
As illustrated in Fig.~\ref{fig:4}, we compare qualitatively with current SOTA methods \cite{PhotoMaker-2023, FlashFace-2024, IP-Adapter-2023, InstantID-2024}. For a fair comparison, we employ the optimal versions of IP-Adapter \cite{IP-Adapter-FaceID-2024} and InstantID \cite{InstantID-FULL}. All methods use 4 reference portraits from the same ID. To facilitate comparison, we only show 1 reference portrait. The results show that our method leverages face landmarks, which offer customizable facial expressions and poses, unlike previous multi-face methods \cite{FlashFace-2024, PhotoMaker-2023} without face location information. Compared to all evaluated methods, \method\space achieves superior ID preservation, enhanced facial controllability, and better aesthetics.






\vspace{-0.5cm}
\paragraph{Quantitative results.}
Tab.~\ref{tab:qualitative} presents the qualitative comparisons on evaluation datasets. To ensure fairness, we inject landmark images via ControlNet \cite{ControlNet-2023} for all methods \cite{PhotoMaker-2023, FlashFace-2024, IP-Adapter-2023, ConsistentID-2024} that lack control information. The results demonstrate that our proposed \method\space outperforms existing methods in ID fidelity ($Sim_{f}$, $CLIP_{f}$). This is attributed to the utilization of the face refiner and HiFi-Net, which facilitate the generation of high-fidelity face features. Moreover, our approach achieves superior results in $Exp$ and $Pose$, benefiting from the guided fusion of multi-face features by face landmarks in HiFi-Net for precise facial control. Furthermore, our method generates high-quality images with clear advantages in FID scores.

\vspace{-0.5cm}
\paragraph{User study.}
To further demonstrate the superiority of \method, we randomly select 20 IDs (5 prompts, 3 seeds) from WebFace-100 and generate results using each method. Twenty users rate the images based on ID fidelity, face controllability, and aesthetic quality. As shown in Fig.~\ref{fig:user_study}, our model consistently receives the highest preference across all metrics.

\vspace{-0.5cm}
\paragraph{Compatibility.}
In Fig.~\ref{fig:application}, we utilize SDXL-based models (including 3D cartoon \cite{3DRenderingStyle2023}, 2D anime \cite{SDXLYamersAnime2024} and pixel art \cite{PixelArtDiffusion2024}) to generate images with different styles. As can be seen from the displayed results, although \method\space focuses on real-life portrait generation, it can still be used to generate images of different styles and achieve high-fidelity ID preservation. This also illustrates the diversity and compatibility of our approach.

Fig.~\ref{fig:ours_instantid} demonstrates the results when \method\space is combined with InstantID. The comparison reveals that InstantID, relying solely on text prompts and face keypoints, fails to control facial expressions precisely. In contrast, our method leverages face landmarks to accurately dictate expressions and poses accurately, achieving superior ID fidelity. Moreover, \method\space is also compatible with InstantID. 


\begin{table}[t!]
\caption{
    Ablation on HiFi-Net.
}
\label{tab:HiFi-Net}
\centering
\begin{adjustbox}{max width=0.98\linewidth}
\begin{tabular}{@{}cccccc@{}}
\toprule
\textbf{HiFi-Net} & \textbf{Condition} & $Sim_{f}\uparrow$ & $CLIP_{f}\uparrow$ & $Exp\downarrow$ & $Pose\downarrow$ \\ \midrule
\ding{55} & \ding{55} & 61.8 & 52.5 & 22.7 & 8.4 \\
\ding{51} & keypoints & 64.5 & 58.1 & 19.8 & 7.2 \\
\ding{51} & \textbf{landmarks} & \textbf{65.2} & \textbf{58.9} & \textbf{18.4} & \textbf{6.8} \\ \bottomrule
\end{tabular}
\end{adjustbox}
\vspace{-0.3cm}
\end{table}

\begin{figure*}[t!]
    \centering
    \begin{minipage}{0.77\linewidth}
        \centering
        \includegraphics[width=\linewidth]{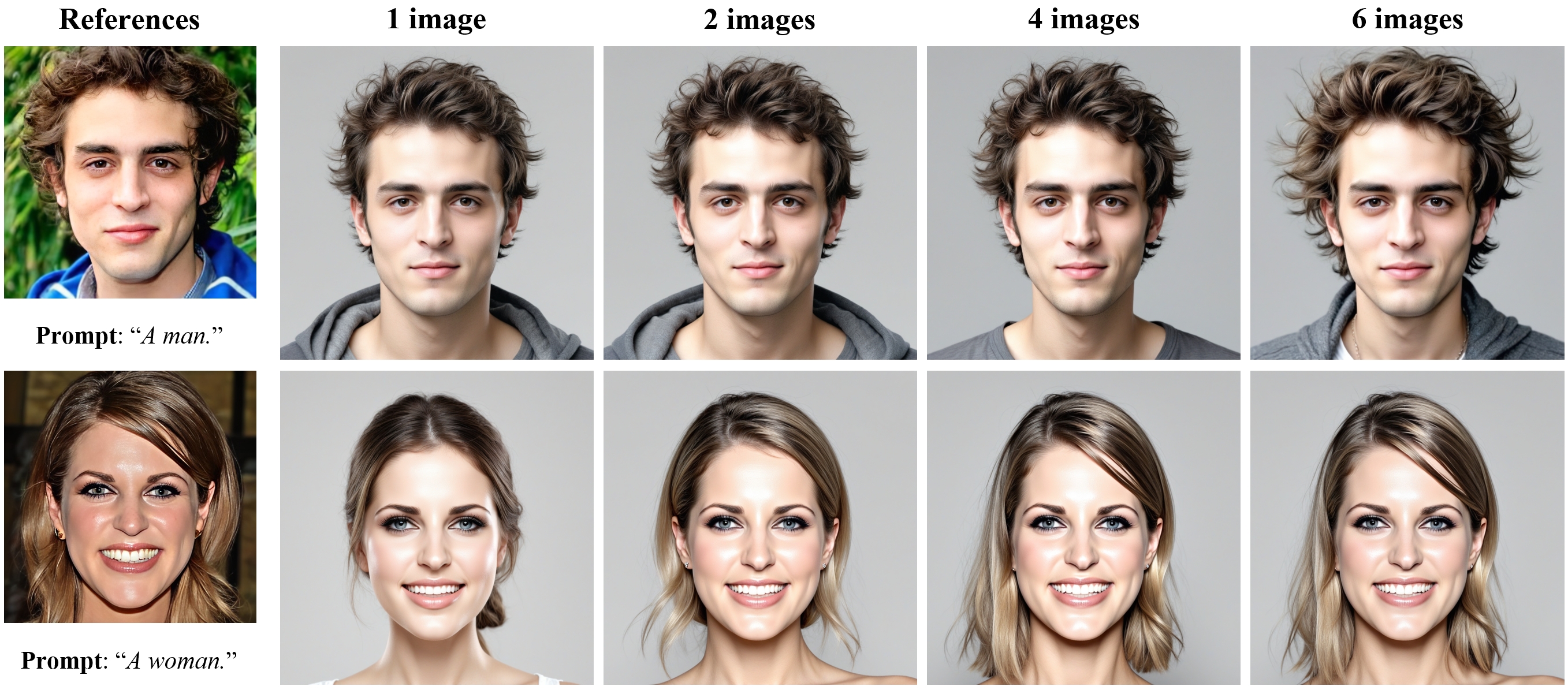}
        \caption{
            \textbf{Impact of the number of reference images.} The reference and target images originating from the same ID. The results indicate that ID fidelity improves with an increasing number of reference images, underscoring the effectiveness of our multi-face fusion strategy.
        }
        \label{fig:n_refs}
    \end{minipage}
    \hspace{0.02\linewidth}
    \begin{minipage}{0.18\linewidth}
    \captionsetup{type=table}
    \caption{
        Ablation on the number of reference images.
    }
    \label{tab:num_ref}
    \vspace{0.3em}
    \scalebox{0.8}{
        \begin{tabular}{@{}ccc@{}}
            \toprule
            \textbf{\textit{N}} & $Sim_{f}\uparrow$ & $CLIP_{f}\uparrow$ \\ 
            \midrule
            1 & 61.9 & 52.3 \\
            2 & 63.8 & 56.7 \\
            3 & 64.6 & 57.1 \\ 
            \textbf{4} & \textbf{65.2} & \textbf{58.9} \\ 
            5 & 65.0 & 58.6 \\
            6 & 64.8 & 58.5 \\
            \bottomrule
        \end{tabular}
    }
    \end{minipage}
\vspace{-0.3cm}
\end{figure*}

\vspace{-0.05cm}
\subsection{Ablation Studies} 
\label{subsec:ablation}
\vspace{-0.05cm}
Following previous works \cite{PhotoMaker-2023, FlashFace-2024}, we shortened the total number of training steps to one-eighth to allow for ablations on each trainable module. All ablations are performed on WebFace-100.

\vspace{-0.3cm}
\paragraph{Impact of face refiner.} 
Tab.~\ref{tab:face_refiner} illustrates the influence of face refiner on maintaining ID similarity. Rows 1 to 3 correspond to local features, global features, and their concatenation are respectively fed into the HiFi-Net. Note that they are all multi-face features. Fig.~\ref{fig:face_refiner} presents the visual results that utilize various facial features. Experimental results show that both local and global features contribute to preserving ID. Utilizing the face refiner enhances ID details in multi-face features, auguring ID fidelity.

\vspace{-0.4cm}
\paragraph{Impact of HiFi-Net.}
In Tab.~\ref{tab:HiFi-Net}, we investigate the impact of HiFi-Net. Row 1 represents HiFi-Net and facial location information is not utilized, instead employing Adapter (trainable cross-attention layers) \cite{IP-Adapter-2023} to inject multi-face features into the UNet. Rows 2 and 3, building on HiFi-Net, leverage face keypoints and landmarks as conditional inputs, respectively. In addition, we visualize generated results under different conditions, as shown in Fig.~\ref{fig:HiFi_Net}. Results indicate that Adapter yields poor outcomes due to the lack of guidance on facial location. Utilizing keypoints within HiFi-Net produces moderate results but lacks control over facial expressions. Conversely, using landmarks to guide multi-face feature fusion within HiFi-Net enhances face fidelity and controllability.

\vspace{-0.4cm}
\paragraph{The number of reference images $N$.}
Tab.~\ref{tab:num_ref} reports on the effect of using different numbers of reference images during inference. Compared with $N=1$, notable improvements are observed when $N=2$. The best results are achieved when $N=4$. In the Fig.~\ref{fig:n_refs}, we visualize the effect of the number of reference images on the generated results. It can be seen that the ID fidelity increases as the number of reference images increases

\vspace{-0.4cm}
\paragraph{Effect of landmark selection.}
Tab.~\ref{tab:landmark} explores the effect of the selected image in landmark generator during inference. Row 1 directly uses target landmarks (ID, pose, expression from target), resulting in lower ID fidelity. Compared to randomly selecting a reference image (row 2), choosing images with the highest face quality score or angle score (row 6) yields better performances. Row 5 is derived by manually selecting high-quality images with face angles close to the target image. In practical applications, allowing user manual selection might be considered. Compared to cumbersome manual selection, row 6 employs face quality and angle scores, improving ID fidelity while maintaining robust facial control. Therefore, row 6 is adopted during inference.




\begin{table}[t!]
\caption{
    Ablation on landmark selection.
}
\label{tab:landmark}
\centering
\scalebox{0.85}{
\begin{tabular}{@{}ccccc@{}}
\toprule
\textbf{Method} & $Sim_{f}\uparrow$ & $CLIP_{f}\uparrow$ & $Exp\downarrow$ & $Pose\downarrow$ \\ \midrule
target & 60.2 & 56.4 & \textbf{18.0} & \textbf{6.6} \\
random & 64.5 & 58.2 & 18.8 & 7.1 \\
quality & 64.8 & 58.7 & 18.2 & 6.9 \\
angle & 64.9 & 58.4 & 18.5 & 6.8 \\
manual & 65.1 & 58.7 & 18.6 & 6.9 \\
\textbf{quality \& angle} & \textbf{65.2} & \textbf{58.9} & 18.2 & 6.8 \\ 
\bottomrule
\end{tabular}
}
\vspace{-0.2cm}
\end{table}



\section{Conclusion}
\vspace{-0.1cm}
This paper presents \method, a high-fidelity ID-preserved framework for zero-shot portrait generation. To achieve this, we design the face refiner and landmark generator to obtain multi-face features and 3D-aware face priors. Next, we develop HiFi-Net leveraging landmark conditioning to effectively guide the fusion of multi-face features, which enhances ID fidelity and face controllability. Finally, we engineer an automated pipeline to collect an ID-based dataset, facilitating the training of \method. Compared with the SOTA methods, our approach can generate more high-quality portraits with customizable facial attributes, while maintaining strong ID fidelity. More discussions can be found in \textbf{\textit{appendix}}.



\clearpage
{
    \small
    \bibliographystyle{cvpr25}
    \bibliography{main}
}

\clearpage
\setcounter{page}{1}
\maketitlesupplementary
\appendix

\section*{Appendix} 

In this appendix, we provide the following details:

\space\space 
(1) A comprehensive statistics of our proposed dataset, presented in Sec.~\ref{sec:dataset}.

\space\space 
(2) A detailed description of the HiFi-Net architecture, featured in Sec.~\ref{sec:HiFi-Net}.

\space\space 
(3) Supplementary experiments, presented in Sec.~\ref{sec:visulization}.

(4) Further discussions, found in Sec.~\ref{sec:discussion}.

\section{Detailed Dataset Statistics}
\label{sec:dataset}

To train our proposed \method, we construct a high-quality ID-based dataset. In this section, we count identities, the number of images, identity cluster size, face area, age and gender distribution, to highlight the diversity of our dataset.

\paragraph{Statistics for identities and images.}
As illustrated in Tab.~\ref{tab:dataset}, the dataset consists of three parts: VGGFace2-HQ \cite{SimSwapPP-VGGFaceHD-2023}, IMDb-Face \cite{dataset-IMDb-Face-2018}, Web. Among them, VGGFace2-HQ is the high-resolution version of VGGFace2 \cite{dataset-VGGFace2-2018}. Its images utilize face alignment; therefore, the face area is relatively large. To improve diversity, we download the source images of IMDb-Face and collect additional data from the Internet (abbreviated as "Web"). After extensive data cleansing and processing, our final dataset consists of 34k IDs and 960k images, averaging 28.2 images per ID, underscoring the diversity of identities.

\begin{table}[h!]
\centering
\caption{
    \textbf{Statistics of identities and images of our dataset}. It consists of three parts: VGGFace2-HD \cite{SimSwapPP-VGGFaceHD-2023}, IMDb-Face \cite{dataset-IMDb-Face-2018}, Web (data we collected from the Internet).
}
\label{tab:dataset}
\begin{adjustbox}{max width=\linewidth}
\begin{tabular}{@{}cccc@{}}
\toprule
\textbf{Dataset }& \textbf{Identities} & \textbf{Images} & \textbf{Images/ID} \\ 
\midrule
VGGFace2-HD & 8k & 440k & 55.0 \\
IMDb-Face & 21k & 380k & 18.1 \\
Web & 5k & 140k & 28.0 \\
\textbf{All} & 34k & 960k & 28.2 \\ 
\bottomrule
\end{tabular}
\end{adjustbox}
\end{table}

\begin{figure}[h]
  \centering
  \includegraphics[width=\linewidth]{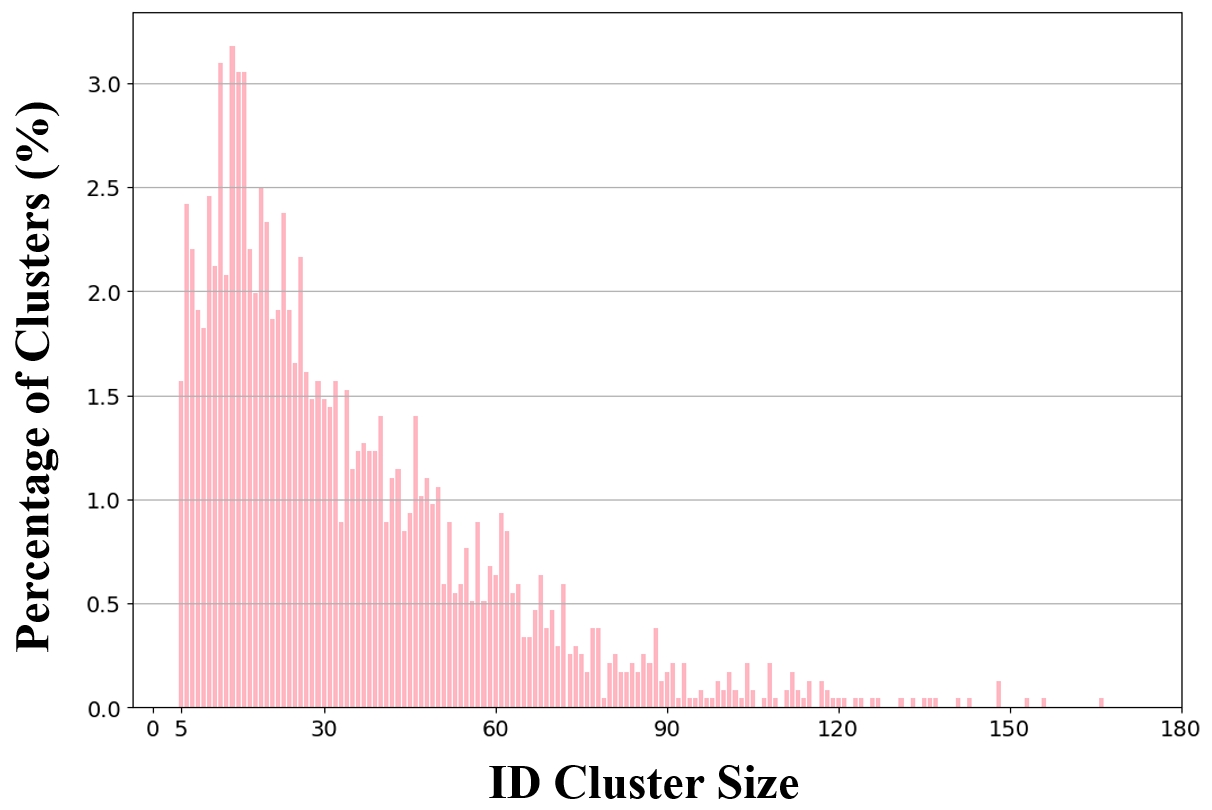}
  \vspace{-0.2cm}
  \caption{
    \textbf{Distribution of ID cluster size.}
  }
  \label{fig:id_cluster}
\end{figure}

\begin{figure}[h!]
  \centering
  \includegraphics[width=0.95\linewidth]{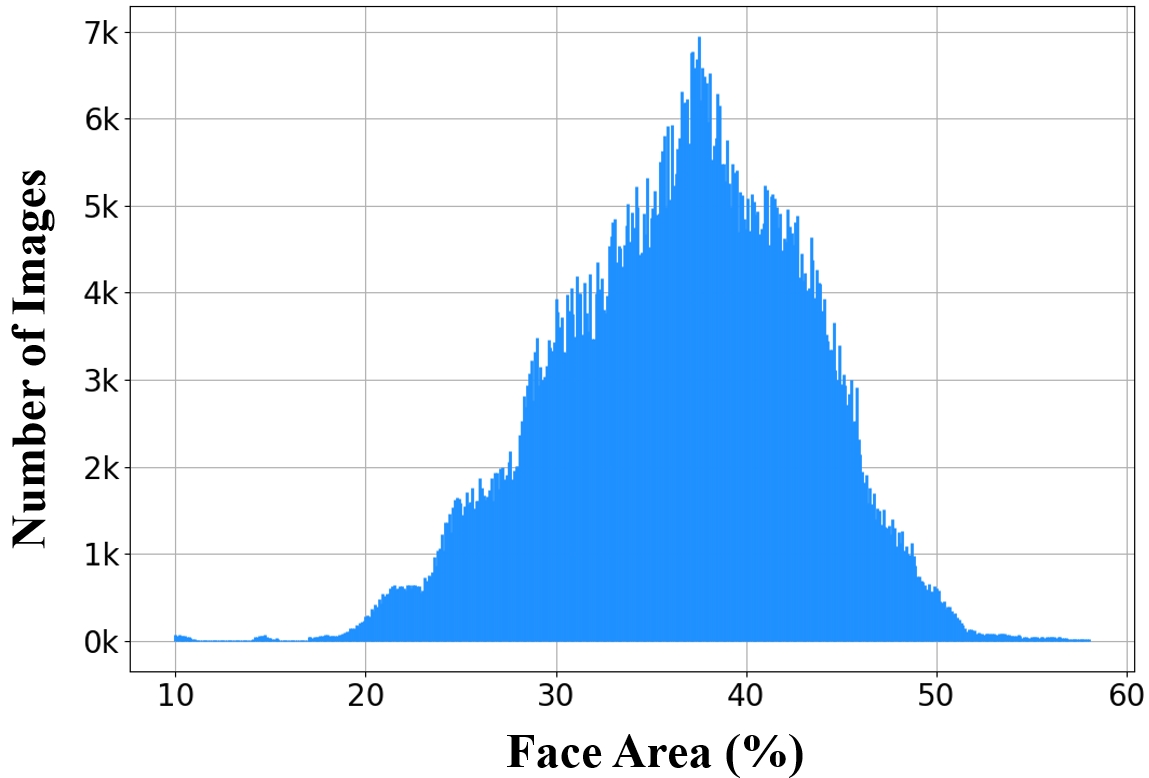}
  \vspace{-0.2cm}
  \caption{
    \textbf{Distribution of face area.}
  }
  \label{fig:face_area}
\end{figure}

\begin{figure}[h!]
  \centering
  \includegraphics[width=0.95\linewidth]{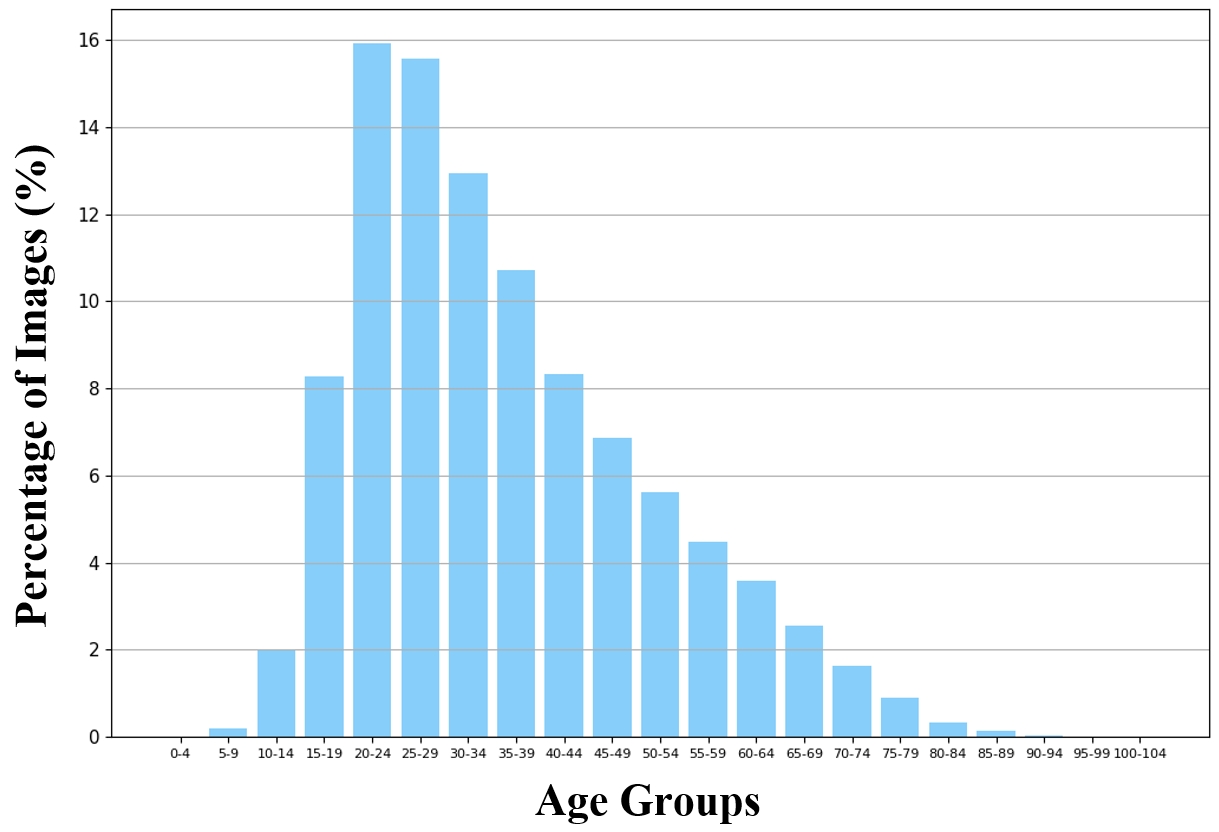}
  \vspace{-0.2cm}
  \caption{
    \textbf{Age distribution.}
  }
  \label{fig:age}
\end{figure}

\begin{figure}[t!]
  \centering
  \includegraphics[width=\linewidth]{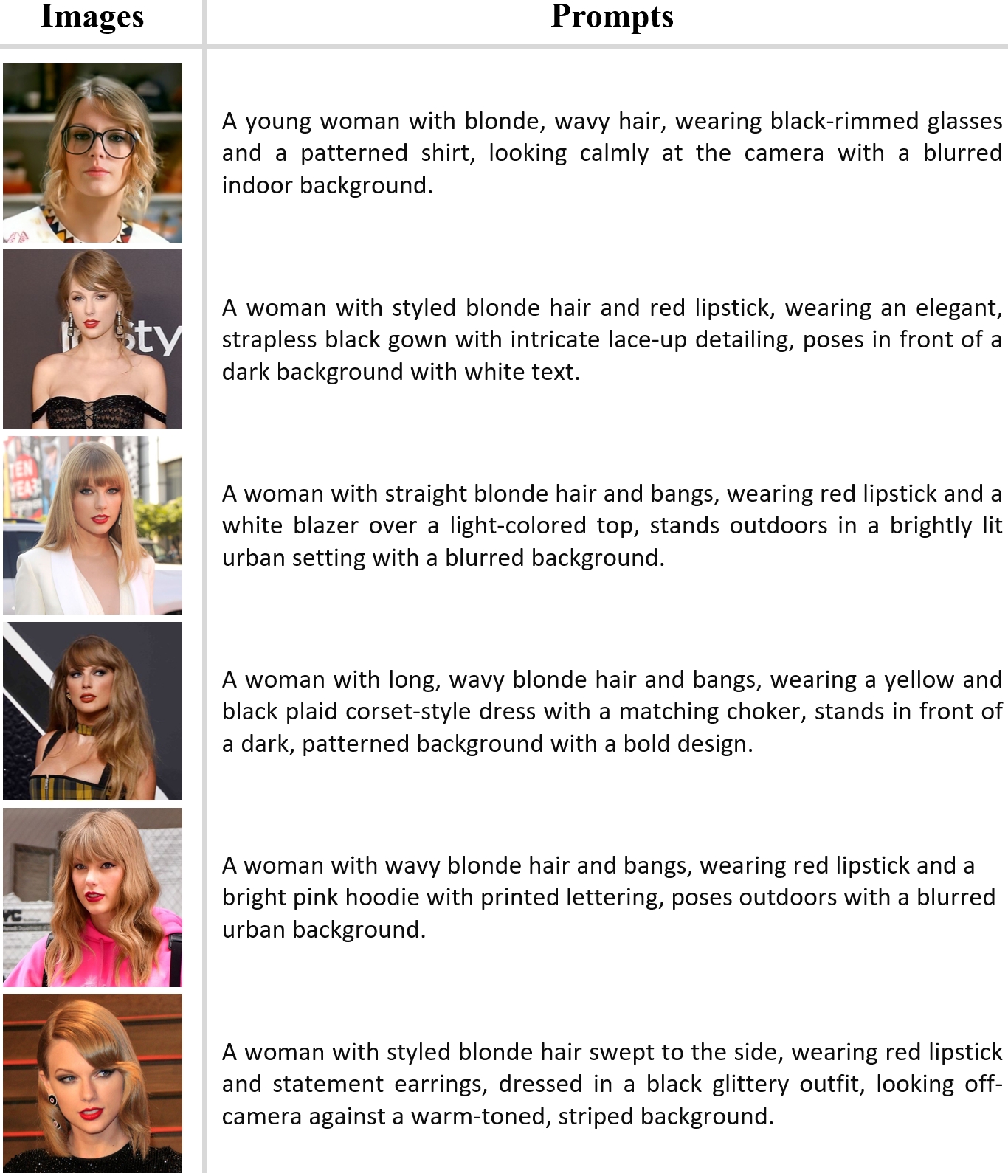}
  \caption{
    \textbf{Some samples from the same ID group.}
  }
  \label{fig:cases}
\end{figure}

\begin{figure}[t!]
  \centering
  \includegraphics[width=\linewidth]{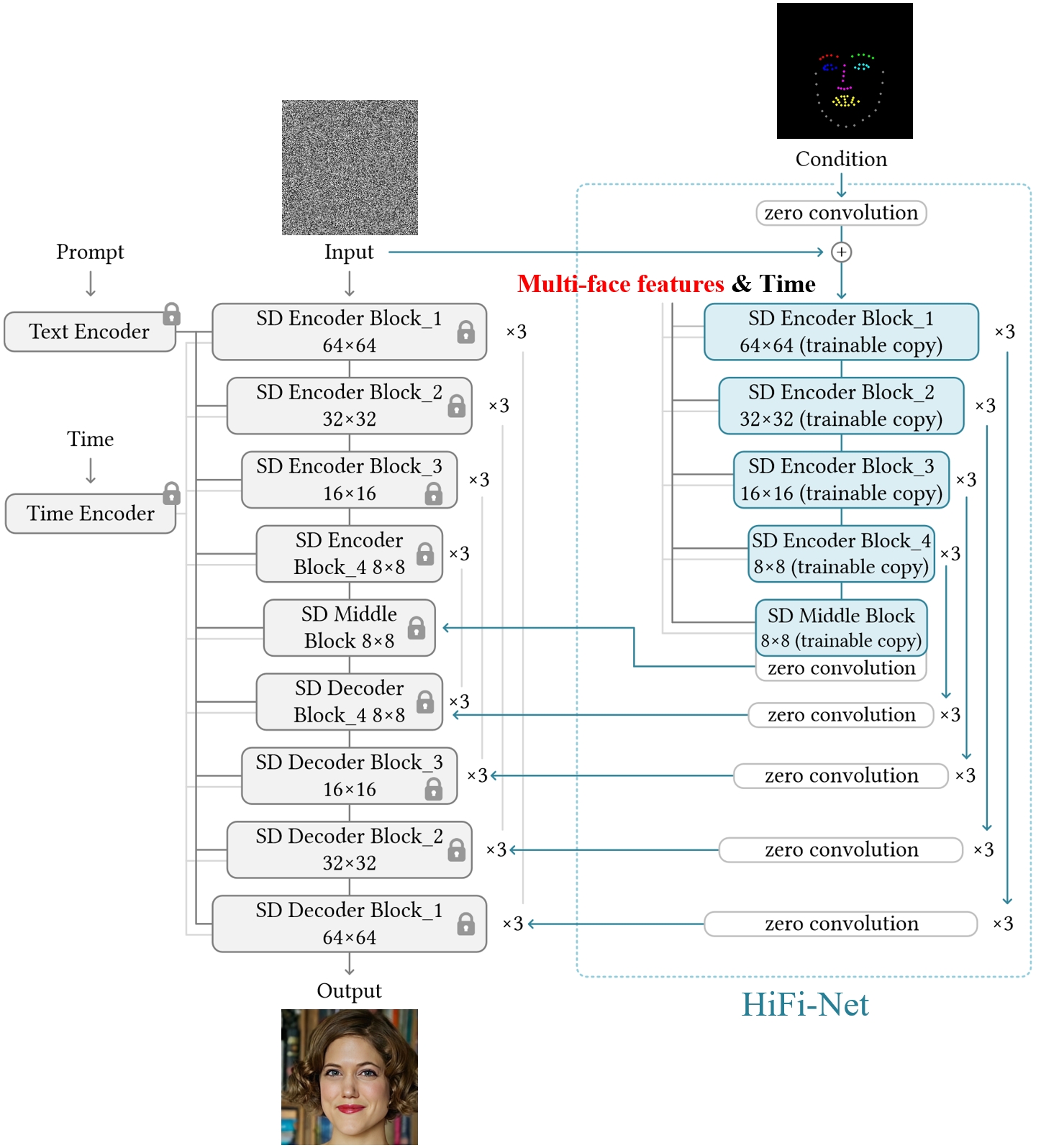}
  \caption{
    \textbf{The architecture of HiFi-Net.} The input condition is the face landmarks, and the input cross-attention condition is multi-face features. The landmarks guide the multi-face features to be effectively aligned and fused in HiFi-Net.
  }
  \label{fig:HiFi-Net_framework}
\end{figure}

\begin{figure}[t!]
  \centering
  \includegraphics[width=0.38\linewidth]{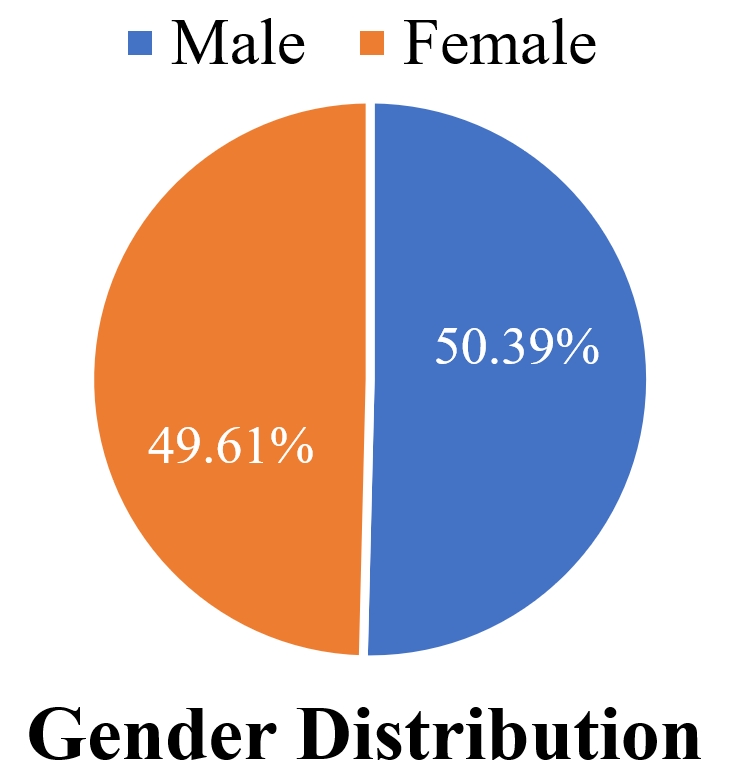}
  \caption{
    \textbf{Gender distribution.}
  }
  \label{fig:gender}
\end{figure}

\paragraph{Identity cluster size.}
Fig.~\ref{fig:id_cluster} presents the distribution of each ID cluster size. Ince the model training requires at least 1 target image and 4 reference images, ID clusters with fewer than five images are excluded.

\paragraph{Face area.}
Fig.~\ref{fig:face_area} displays the distribution of face area proportions within the dataset, where the face area is defined by the bounding box area predicted by RetinaFace \cite{RetinaFace-2020}. The spacing on the x-axis is set at 0.1. We observe that images with small faces often feature blurry and distorted faces; therefore, we filter out images where the face area is less than 10\%.

\paragraph{Age and gender distribution.}
Fig.~\ref{fig:age} and \ref{fig:gender} outline the distribution of age groups and genders, respectively. Here, age is estimated by Facelib \cite{ArcFace-2019}. For gender, both IMDb-Face and Web contain names, and we can retrieve gender by name. For images where gender could not be directly retrieved, we use Facelib to obtain their gender.

\paragraph{Dataset sample.}
In Fig.~\ref{fig:cases}, we present several samples from the same ID group in the dataset.


\section{HiFi-Net}
\label{sec:HiFi-Net}
The architecture of HiFi-Net is depicted in Fig.~\ref{fig:HiFi-Net_framework}. We made two critical modifications compared to ControlNet \cite{ControlNet-2023}: 

1. Face landmarks are used as input conditions to guide multi-face fusion, which enables more precise control over facial expressions and poses, thereby enhancing ID fidelity as illustrated in Fig.~\ref{fig:ours_instantid}~and~\ref{fig:HiFi_Net}.

2. The cross-attention condition is multi-face features derived from the face refiner. Multi-face features capture more facial details than local and global features, as shown in Fig.~\ref{fig:face_refiner}~and~\ref{fig:HiFi_Net}. In addition, compared with single-face features, multi-face features contain richer ID information, as shown in Fig.~\ref{fig:n_refs}.

In summary, we employ HiFi-Net for effective multi-face fusion.

\begin{figure*}[ht!]
  \centering
  \includegraphics[width=\linewidth]{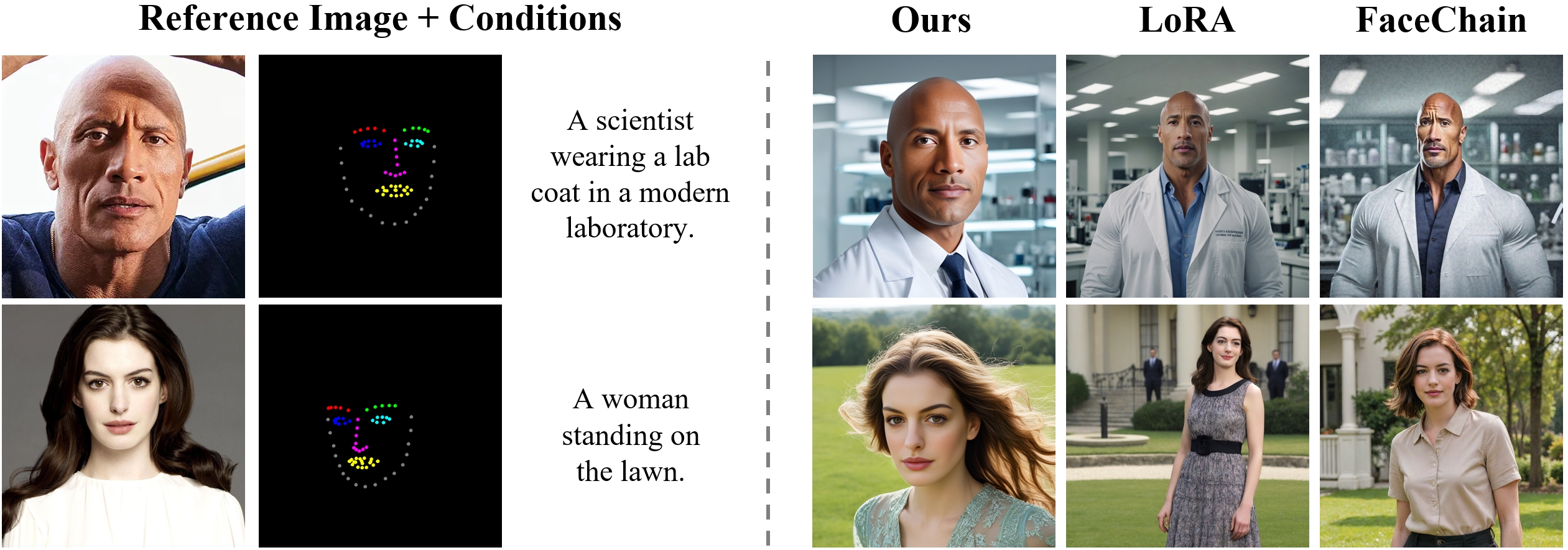}
  \caption{
    \textbf{Comparison with LoRA-based methods.} Our method exhibits higher ID fidelity.
  }
  \label{fig:lora}
\end{figure*}

\begin{figure*}[h]
  \centering
  \includegraphics[width=\linewidth]{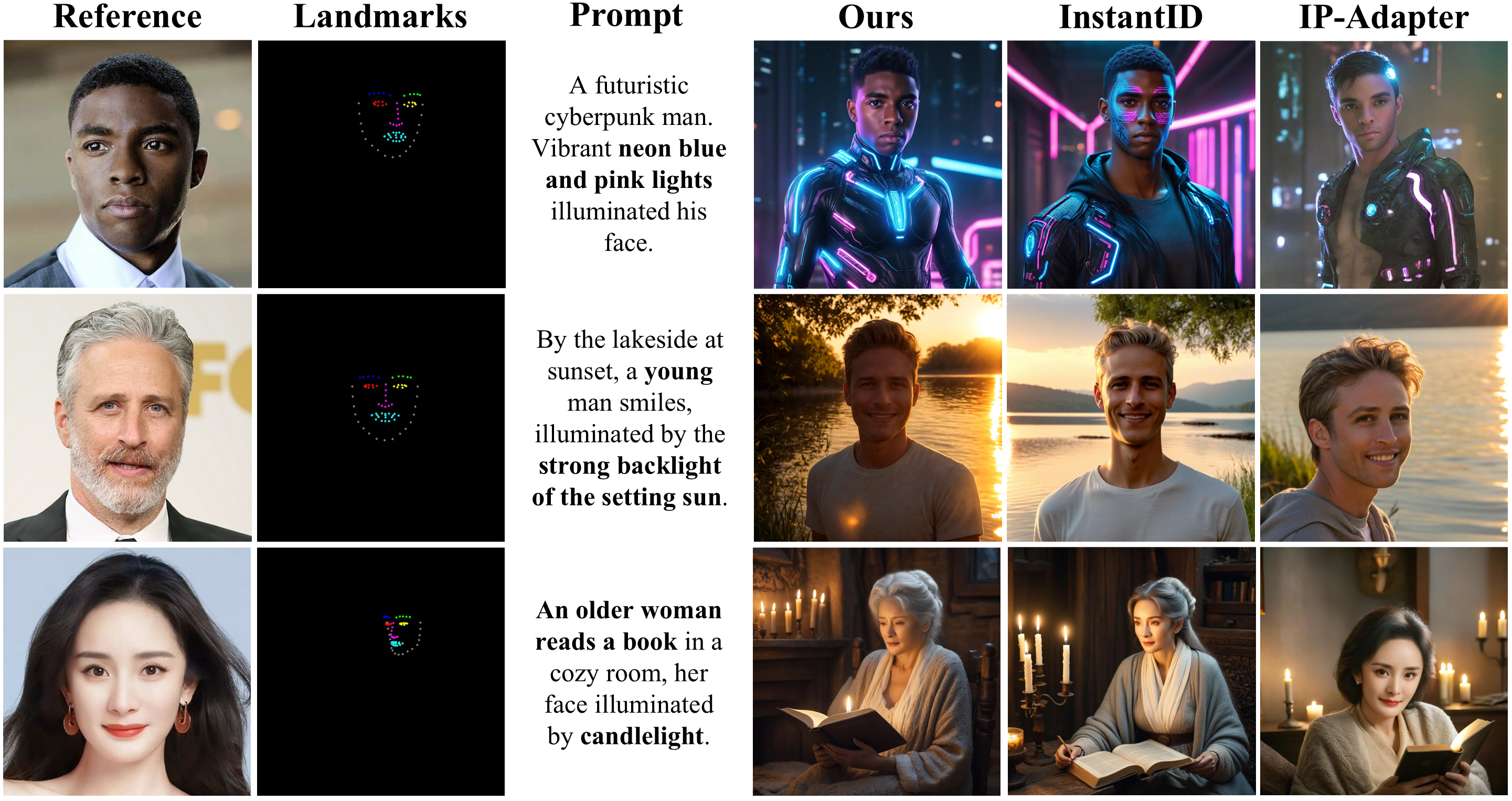}
  \caption{
    \method\space generates \textbf{lighting} that is more consistent with prompt. Zoom in the image for a better visual experience.
  }
  \label{fig:lighting}
\end{figure*}
\begin{figure*}[h]
  \centering
  \includegraphics[width=\linewidth]{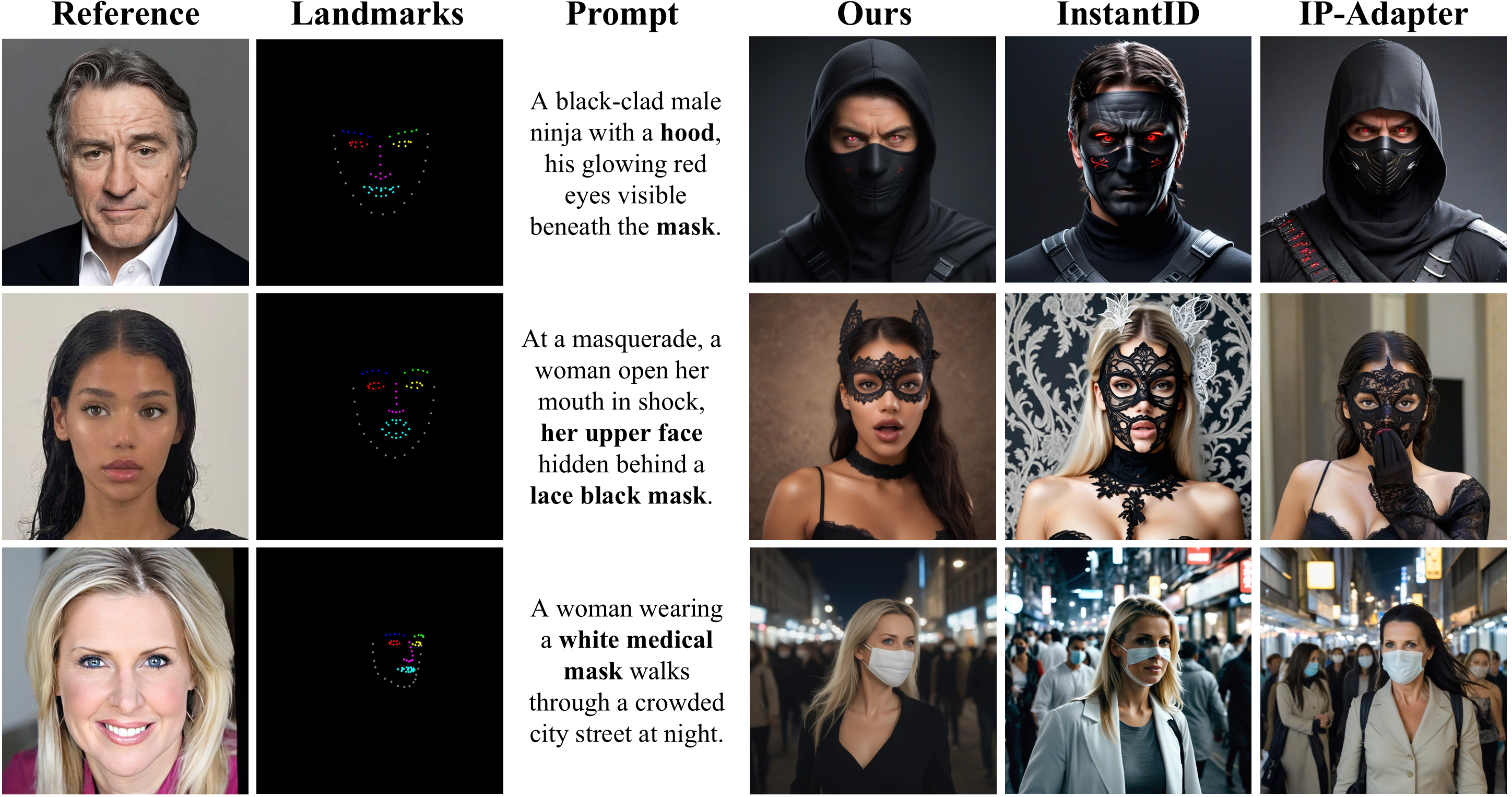}
  \caption{
    Our method generates more natural \textbf{facial occlusions}.
  }
  \label{fig:occlusion}
\end{figure*}
\begin{figure*}[h!]
  \centering
  \includegraphics[width=\linewidth]{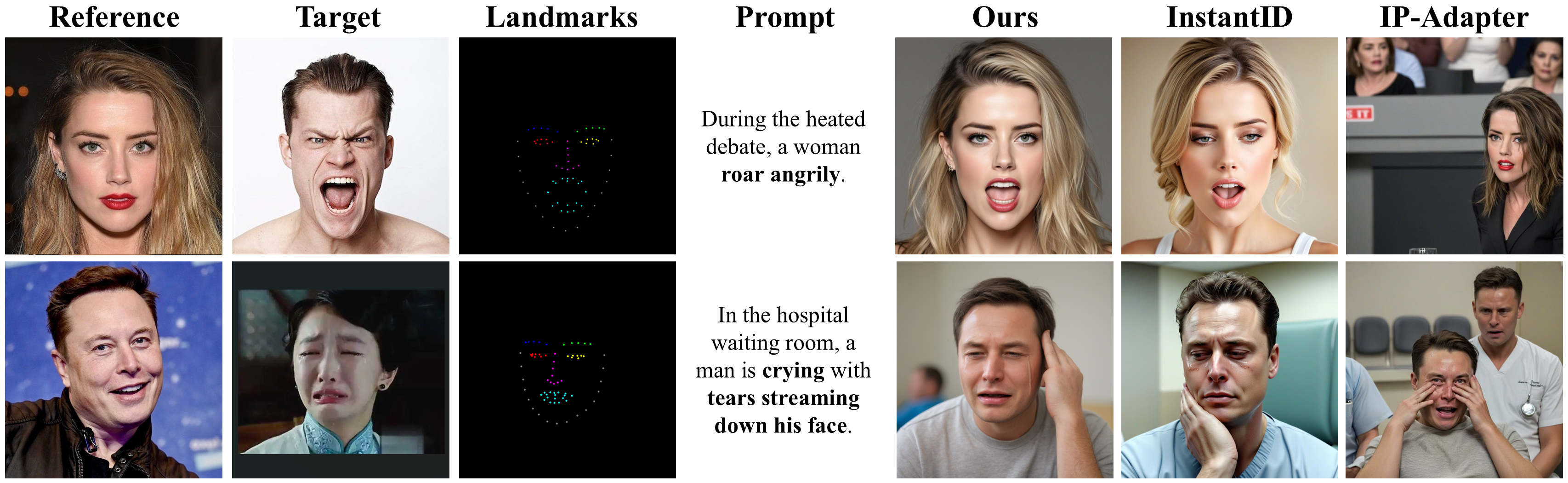}
  \caption{
    Our method effectively remains \textbf{extreme expressions}.
  }
  \label{fig:expr}
\end{figure*}
\begin{figure*}[h!]
  \centering
  \includegraphics[width=\linewidth]{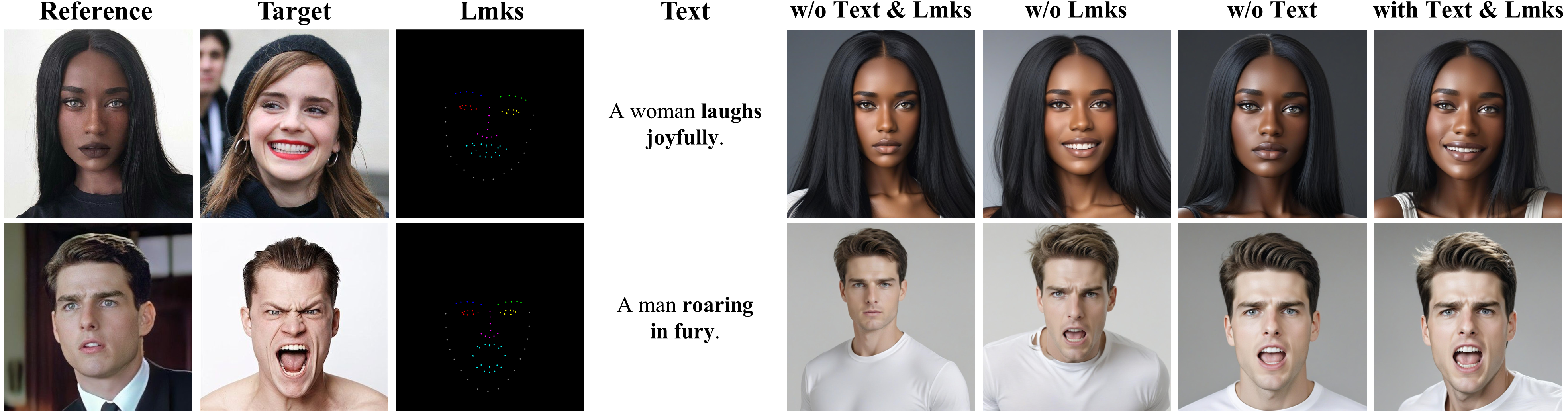}
  \caption{
    Using text prompt improves the alignment between the target and  generated expressions.
  }
  \label{fig:txt_img}
\end{figure*}

\section{Supplementary experiments}
\label{sec:visulization}

\subsection{Comparison with LoRA-based methods}
Fig.~\ref{fig:lora} shows the qualitative comparison between \method\space and LoRA-based methods \cite{LoRA-2021, FaceChain-2023}. Our method demonstrates higher ID fidelity and precisely controls the face expression and pose.

\subsection{More challenging cases}
Figs.~\ref{fig:lighting}–\ref{fig:expr} illustrates varying lighting, facial occlusion, and extreme expressions. Fig.~\ref{fig:lighting} shows that \method\space generates text-consistent lighting from multiple angles. Fig.~\ref{fig:occlusion} demonstrates more natural facial occlusions. Fig.~\ref{fig:expr} confirms that \method\space effectively remains extreme expressions.

\subsection{Input condition analysis}
Fig.~\ref{fig:txt_img} reports results under different conditions. Col.~5 without text prompts or landmarks. Col.~6 uses text only. Comparing Col.~7 (uses landmarks only) and Col.~8 (uses text and landmarks) demonstrates that using text prompt improves the alignment between target and generated expressions.



\section{Further Discussion}
\label{sec:discussion}

\subsection{Limitations and future works}
The primary experiments in this study required approximately 1500 GPU hours on A800 GPUs. Therefore, we do not have enough computing power to explore some hyperparameter settings. Future research should focus on developing more straightforward and controlled models, rather than expending extensive efforts on hyperparameter tuning. Innovations such as a streamlined ControlNet-based method, ControlNext \cite{ControlNext-2024}, or faster denoiser \cite{SDXL-Lightning-2024} could be beneficial. Additionally, experimenting with recent models like SD3 \cite{SD3-2024} or FLUX \cite{FLUX-1-Kontext} may prove fruitful.

\subsection{Broader impacts}
This work contributes a high-fidelity framework for zero-shot ID-preserved generation to the open-source community, capable of customizing facial expressions and poses. Depending on the application context, this may have positive and negative impacts. On the one hand, our multi-face fusion strategy could advance the development of open-source ID-preservation models and use them in practical applications. On the other hand, the high fidelity of generated images could be misused for facial fraud.

\subsection{Ethical considerations}
To facilitate a more accurate understanding of human anatomy, SDXL utilizes a limited number of nude images for training. In our experiments, if clothing is not specified in the text prompts, there is a small probability that the generated portraits may be nude. Consequently, we recommend activating NSFW detection to minimize this issue. Additionally, the generative results have observed no other unethical or harmful behaviors. Finally, it is imperative to note that all data and models in this paper are intended strictly for research purposes and must not be used commercially.

\end{document}